\documentclass[acmlarge]{acmart}

\AtBeginDocument{%
  }

\setcopyright{acmlicensed}
\copyrightyear{2024}
\acmYear{2024}
\acmDOI{XXXXXXX.XXXXXXX}

\acmJournal{CSUR}
\acmVolume{37}
\acmNumber{4}
\acmArticle{111}
\usepackage{multirow, multicol}
\usepackage{subfigure}
\usepackage{hyperref}
\usepackage{forest}
\usetikzlibrary{decorations.pathreplacing}
\usepackage[graphicx]{realboxes}
\usepackage{tikz}
\usepackage{array}
\usepackage{multirow, hyperref, bookmark}
\usepackage{booktabs}
\usepackage{tcolorbox}
\usepackage{xcolor}

\usepackage{makecell, mdframed, framed}
\usepackage{tabularx}
\usepackage{colortbl}
\usepackage{longtable, appendix}

\begin{document}

\title{Exploring the Landscape of Text-to-SQL with Large Language Models: Progresses, Challenges and Opportunities}

\author{Yiming Huang}
\email{24b951042@stu.hit.edu.cn}
\orcid{0009-0005-5003-2427}
\affiliation{%
  \institution{Harbin Institute of Technology (Shenzhen)}
  \city{Shenzhen}
  \country{China}
}

\author{Jiyu Guo}
\email{220110126@stu.hit.edu.cn}
\orcid{0009-0001-7055-0767}
\affiliation{%
  \institution{Harbin Institute of Technology (Shenzhen)}
  \city{Shenzhen}
  \country{China}
}

\author{Wenxin Mao}
\email{23s051015@stu.hit.edu.cn}
\affiliation{%
  \institution{Harbin Institute of Technology (Shenzhen)}
  \city{Shenzhen}
  \country{China}
}

\author{Cuiyun Gao}
\email{gaocuiyun@hit.edu.cn}
\orcid{0000-0003-4774-2434}
\affiliation{%
  \institution{Harbin Institute of Technology (Shenzhen), Peng Cheng Laboratory}
  \city{Shenzhen}
  \country{China}
}

\author{Peiyi Han}
\email{hanpeiyi@hit.edu.cn}
\affiliation{%
 \institution{Harbin Institute of Technology (Shenzhen), Peng Cheng Laboratory}
 \city{Shenzhen}
 \country{China}}
 
\author{Chuanyi Liu}
\authornote{Corresponding author}
\email{liuchuanyi@hit.edu.cn}
\affiliation{%
 \institution{Harbin Institute of Technology (Shenzhen), Peng Cheng Laboratory}
 \city{Shenzhen}
 \country{China}}

\author{Qing Ling}
\email{lingqing556@mail.sysu.edu.cn}
\affiliation{%
 \institution{Sun Yat-sen University, Great Bay University}
 \city{Guangzhou, Dongguan}
 \country{China}}

\renewcommand{\shortauthors}{Huang et al.}

\begin{abstract}
Converting natural language (NL) questions into SQL queries, referred to as \textbf{Text-to-SQL}, has emerged as a pivotal technology for facilitating access to relational databases, especially for users without SQL knowledge. Recent progress in large language models (LLMs) has markedly propelled the field of natural language processing (NLP), opening new avenues to improve text-to-SQL systems. This study presents a systematic review of LLM-based text-to-SQL, focusing on four key aspects: (1) an analysis of the research trends in LLM-based text-to-SQL; (2) an in-depth analysis of existing LLM-based text-to-SQL techniques from diverse perspectives; (3) summarization of existing
text-to-SQL datasets and evaluation metrics; and (4) discussion on potential obstacles and avenues for future exploration in this domain. This survey seeks to furnish researchers with an in-depth understanding of LLM-based text-to-SQL, sparking new innovations and advancements in this field.
\end{abstract}




\maketitle

\section{Introduction}
\begin{figure*}[htbp]
\centering 
\includegraphics[width=1.0\columnwidth]{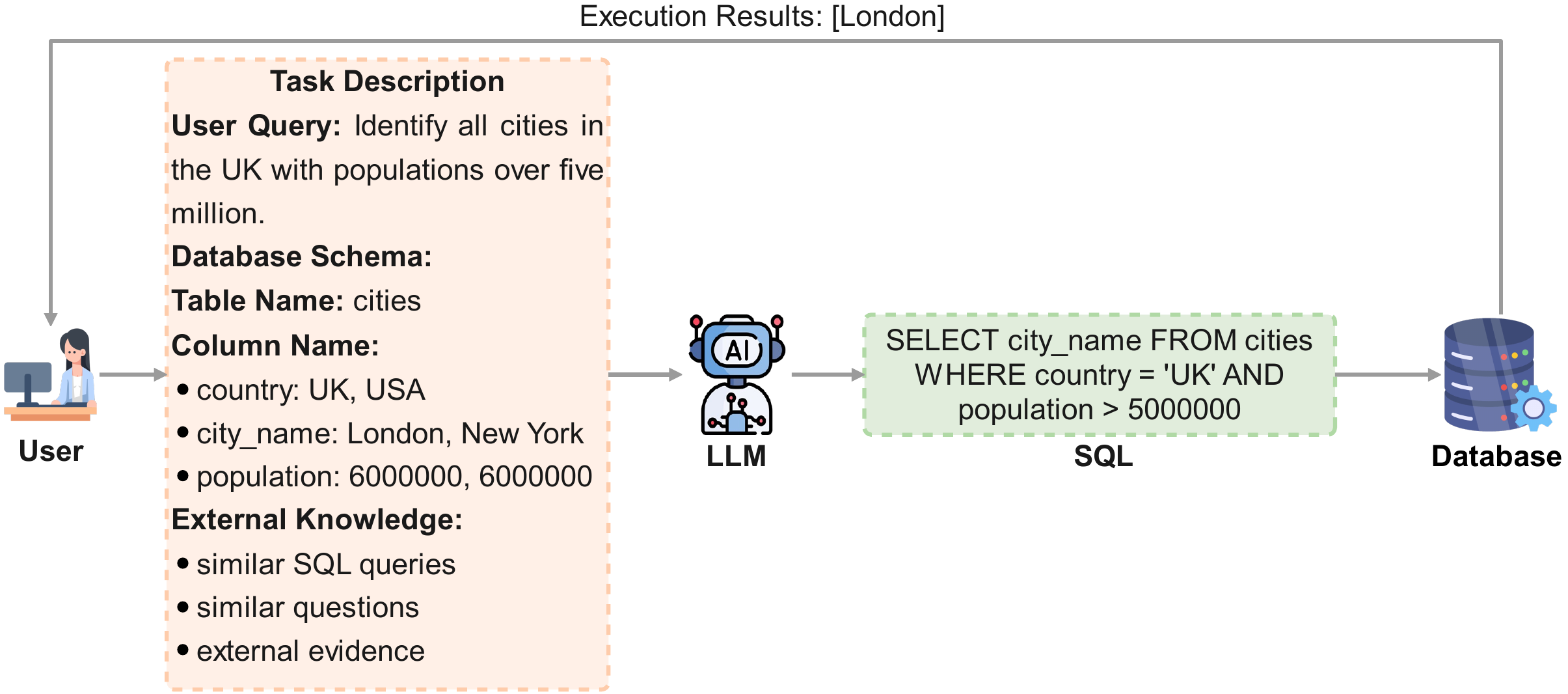}
\caption{An illustrative example of using LLMs for text-to-SQL tasks. Specifically, a user requests LLMs to transform a provided question into its corresponding SQL. The produced SQL is then executed in the given database, and the resulting execution outcomes are subsequently transmitted back to the user.}
\label{fig1}
\end{figure*}

The utilization of SQL queries has significantly improved the efficiency of extracting data from multiple databases. These data have been effectively applied in various essential domains, such as business intelligence \cite{mohammed2024towards} and healthcare analytics \cite{mendhe2024ai}. While technical professionals are adept at handling SQL queries, natural language interfaces to databases (NLIDBs) have empowered non-technical users to extract information from structured databases seamlessly \cite{deng2022recent}. This accessibility has markedly propelled the advancement of text-to-SQL systems, which automatically transform natural language (NL) queries into valid SQL. To illustrate this, we consider a simple database depicted in Fig. \ref{fig1}, which includes a table called ``cities'' with columns named ``country'', ``city\_name'', and ``population''. By using text-to-SQL systems, a user query like ``Identify all cities in the UK with populations over five million'' would be transformed into the following SQL: ``SELECT city\_name FROM cities WHERE country = `UK' AND population > 5000000''. The produced SQL is subsequently executed within the given database, and the resulting data, such as ``[London]'', is fed back to the user. This entire process simplifies access to information for users who are not familiar with SQL knowledge.

In recent years, the field of natural language processing (NLP) has witnessed a remarkable breakthrough with the emergence of large language models (LLMs) \cite{achiam2023gpt, guo2024deepseek}. These models, equipped with their unprecedented capacity to process and produce human-like context, have garnered considerable attention \cite{chang2024survey, yang2024harnessing}. As LLMs continue to evolve, new capabilities begin to emerge, such as zero-shot learning \cite{wei2021finetuned}, few-shot learning \cite{song2023comprehensive}, and instruction-following \cite{wu2024language}. In light of these abilities, text-to-SQL approaches based on LLMs have gained prominence, particularly with the advent of \textit{In-Context Learning-based (ICL-based)} \cite{pourreza2024din} and \textit{Fine-Tuning-based (FT-based)} techniques \cite{li2024codes}. Consequently, it is essential for researchers to have a systematic comprehension of the key methodologies, challenges, and future directions of LLM-based text-to-SQL. Considering such significance, we conducted a systematic overview of LLM-based SQL generation approaches, analyzing 92 relevant articles published from April 2022 to October 2024. In addition, we selected open-source datasets and evaluation metrics regarding text-to-SQL from 2017 to October 2024. By examining the development trends in LLM-based text-to-SQL and investigating a range of research works from diverse perspectives, we aimed to outline potential challenges and avenues for future investigations. In summary, this literature review offers the following contributions.

\begin{itemize}
    \item \textbf{Research Trend Analysis of the Surveyed Articles.} We conduct an analysis of the surveyed articles to identify research trends according to publication dates, publication venues, and main types of contributions.

    \item \textbf{Overview of the Surveyed Articles.} We categorize LLM-based text-to-SQL studies into three groups based on their research goals (i.e., methodologies, datasets, and evaluation metrics). Each category is subsequently introduced in a sequential manner to provide a systematic overview of the surveyed papers.

    \item \textbf{New Taxonomy for LLM-based Approaches.} We present a new taxonomy for SQL generation methods using LLMs, categorizing them into four main paradigms: \textit{Pre-Processing}, \textit{In-Context Learning}, \textit{Fine-Tuning}, and \textit{Post-Processing}. Each area is further divided into subcategories based on specific model designs, offering a systematic review of the current innovations in this field. 

    \item \textbf{Overview of Datasets and Evaluation Metrics.} We present an overview of existing datasets and evaluation metrics in text-to-SQL tasks.

    \item \textbf{Discussion of Current Challenges and Future Directions.} Through the analysis of the surveyed articles, we identify several significant obstacles in current research. Additionally, key future directions are further discussed, offering guidance for developing more robust, efficient, and reliable systems. 
    
\end{itemize}

 The remainder of this article is organized as follows: Section \ref{background} introduces the evolution of text-to-SQL models, emphasizing the reasons for using LLMs in SQL generation. Section \ref{method} outlines the methodology used for our systematic literature review. Section \ref{RQ1} examines research trends in LLM-based text-to-SQL. Section \ref{RQ2} explores existing methodologies, categorizing them into pre-processing, in-context learning, fine-tuning, and post-processing paradigms. Section \ref{RQ3} provides a review of existing datasets and evaluation metrics for text-to-SQL. Section \ref{RQ4} discusses current challenges and potential future research directions in text-to-SQL. Section \ref{threat} analyzes threats to the validity of our study, and Section \ref{conclusion} provides a summary of this survey. We hope this review can offer a clear landscape of current advancements and inspire future exploration in LLM-based SQL generation.
 
\section{Background}\label{background}
This section begins with an overview of the evolution of text-to-SQL solutions. Subsequently, key reasons for employing LLMs in current text-to-SQL systems are further discussed. 

\subsection{Model Evolutionary Process}
The text-to-SQL field has evolved significantly, progressing from rule-based systems to neural network-based approaches, and further incorporating pre-trained language models (PLMs) and large language models (LLMs) more recently. These advancements have substantially enhanced the performance of text-to-SQL systems.

\subsubsection{Rule-based Approaches}
Initial SQL generation systems primarily depended on rule-based approaches \cite{yu2020grappa}, which offer the following advantages: (1) \textbf{High Interpretability}: These systems use handcrafted rules, making the translation process transparent and easily understandable. (2) \textbf{Strong Control}: By defining manual translation rules, these systems can ensure consistent and predictable outputs. (3) \textbf{Minimal Data Requirement}: Extensive feature engineering aids in the text-to-SQL process with reduced dependence on large amounts of training data. However, these approaches also present drawbacks: (1) \textbf{Poor Generalization and Scalability}: These systems struggle to handle complex and diverse queries, restricting their practical use in broader text-to-SQL applications. (2) \textbf{High Manual Cost}: The domain-specific nature of rule-based systems demands deep expertise in database schemas to create and update rules, leading to time-consuming and labor-intensive development.

\subsubsection{Neural Network-based Approaches}
The limitations of rule-based methods led to a shift toward neural network-based approaches, which have aroused widespread attention due to deep learning (DL) advancements. Specifically, DL models like Transformers \cite{tyukin2024attention} have been adopted to convert NL inputs into SQL queries \cite{choi2021ryansql}. Meanwhile, graph neural networks (GNNs) have been integrated to represent schema dependencies and to capture relationships among database components \cite{hui2021improving}. Notably, these approaches provide several advantages: (1) \textbf{Enhanced Learning Capabilities}: These models can learn sophisticated mappings from large annotated datasets, improving their ability to process diverse NL inputs. (2) \textbf{Automatic Feature Extraction}: Neural networks can identify relevant features from NL queries and database schemas without extensive manual feature engineering \cite{qin2022survey}. (3) \textbf{Improved Generalization}: With sufficient training data, these models tend to generalize better to new datasets compared to rule-based approaches. However, some disadvantages can also be noticed: (1) \textbf{Intensive Data Requirement}: They necessitate substantial amounts of labeled data for training, which can be costly and time-consuming to acquire. (2) \textbf{Overfitting Risk}: Without adequate regularization techniques, these models may overfit to  training data, potentially leading to poor performance on unseen examples.

\subsubsection{PLM-based Approaches}
Following the conventional neural network-based approaches, pre-trained language models (PLMs) have proven effective in SQL generation tasks, capitalizing on their extensive linguistic knowledge gained during pre-training. Early PLMs like BERT \cite{kenton2019bert} were fine-tuned on text-to-SQL datasets to harness their semantic understanding \cite{yin2020tabert}. By incorporating schema information, PLMs further enhanced their ability to represent database structures, thereby improving accuracy in SQL generation \cite{li2023resdsql}. Notably, PLM-based methods offer several strengths: (1) \textbf{Rich Context Comprehension}: PLMs excel at capturing complex semantic information, making them particularly suitable for tasks that require deep linguistic understanding. (2) \textbf{High Scalability}: By fine-tuning on task-specific datasets, PLMs can effectively apply their pre-trained knowledge to text-to-SQL, which can adapt well to diverse inputs and new contexts. However, these approaches also present several drawbacks: (1) \textbf{Specialized Fine-Tuning}: Fine-tuning for specific tasks often requires model adjustments to avoid negative transfer, which may degrade the model's performance. (2) \textbf{High Dependence on Data}: The performance of PLMs is highly affected by the quantity and quality of task-specific data, making meticulous data preparation an indispensable step.

\subsubsection{LLM-based Approaches}
Currently, large language models (LLMs) have garnered considerable attention in SQL generation owing to their powerful text generation capabilities. Researchers are actively exploring ways to harness LLMs' extensive knowledge and reasoning abilities for SQL generation. For instance, prompt engineering techniques have been developed to effectively guide LLMs in translating NL inputs into accurate SQL queries \cite{pourreza2024din}. Simultaneously, fine-tuning approaches have been employed to enhance the performance of open-source LLMs in SQL generation \cite{li2024codes}. Despite their potential, these approaches commonly face several challenges: (1) \textbf{Model Hallucination}: LLMs occasionally produce information that appears accurate but is factually incorrect, which poses a risk in applications that require high reliability \cite{ji2023survey, farquhar2024detecting}. (2) \textbf{Uncontrollable Results}: Due to their inherent probabilistic nature, LLMs can sometimes produce inconsistent outputs, making it challenging to maintain predictability in their responses \cite{zhang2023survey}. (3) \textbf{High Resource Demands}: The computational and energy requirements for employing open-source LLMs are substantial, especially when using supervised fine-tuning (SFT) for SQL generation \cite{kukreja2024literature}.

\subsection{Reasons for Using LLM-based Text-to-SQL Methods}
Despite the aforementioned limitations, several factors driving the widespread adoption of LLMs in current text-to-SQL systems can be elucidated as follows.

(1) \textbf{Robustness to Linguistic Variability.} NL queries often vary significantly in linguistic structure and complexity. Hopefully, LLMs demonstrate a strong resilience to this variability owing to their extensive pre-training across diverse language patterns. This robustness empowers LLMs to effectively tackle a broader range of query types and complexities for SQL generation. This is particularly advantageous in scenarios where users lack SQL expertise but still need to interact with databases seamlessly.

(2) \textbf{Enhanced Domain Generalization Capability.} LLMs show strong adaptability across different domains, especially through effective prompt engineering techniques. This generalization minimizes the need for retraining, enabling quicker deployment across diverse applications. For instance, in fields like healthcare and finance, LLMs can be leveraged to understand specific terminologies with minimal modifications, streamlining the customization of text-to-SQL systems across specialized domains.

(3) \textbf{Next-Generation Trend.} The ongoing advancements in LLMs are expected to drive continuous innovations in text-to-SQL methods, such as model scaling, advanced prompting techniques, higher-quality datasets, and task-specific fine-tuning approaches. These advancements position LLMs as central to next-generation text-to-SQL systems, bridging the gap between NL and SQL with growing accuracy and efficiency.

\section{Methodology} \label{method}
This section elucidates the approach employed in our investigation of studies on LLM-driven text-to-SQL. We first outline research questions guiding this survey, followed by an introduction of the procedure for retrieving and selecting pertinent articles. Finally, we describe the extracted data items to address each research question.

\subsection{Research Questions}
In this survey, we anticipate to present detailed insights and evidence on current text-to-SQL research by retrieving and classifying existing studies. To guide this effort, we propose the following research questions (RQs):

\begin{itemize}
    \item \textbf{RQ1: What are the research trends in LLM-based text-to-SQL studies?}
    \begin{itemize}
        \item \textit{RQ1.1: What is the trend of studies over time?}
        \item \textit{RQ1.2: What is the distribution of publication venues?}
        \item \textit{RQ1.3: What is the distribution of primary types of contributions?}
    \end{itemize}
    \item \textbf{RQ2: What is the current state of research on LLM-based text-to-SQL techniques?}
    \begin{itemize}
        \item \textit{RQ2.1: What methodologies have been explored in the pre-processing paradigm for LLM-based text-to-SQL?}
        \item \textit{RQ2.2: What methodologies have been explored in the ICL paradigm for LLM-based text-to-SQL?}
        \item \textit{RQ2.3: What methodologies have been explored in the FT paradigm for LLM-based text-to-SQL?}
        \item \textit{RQ2.4: What methodologies have been explored in the post-processing paradigm for LLM-based text-to-SQL?}
    \end{itemize}
    \item \textbf{RQ3: What are the characteristics of text-to-SQL datasets and evaluation metrics?}
    \begin{itemize}
        \item \textit{RQ3.1: What are the existing datasets in text-to-SQL tasks?}
        \item \textit{RQ3.2: What are the existing evaluation metrics in text-to-SQL tasks?}
    \end{itemize}
    \item \textbf{RQ4: What are the challenges and future directions for LLM-based text-to-SQL?}
    \begin{itemize}
        \item \textit{RQ4.1: What are the key unresolved challenges in the surveyed studies?}
        \item \textit{RQ4.2: What are the potential future directions for advancing LLM-based text-to-SQL systems?}
    \end{itemize}
\end{itemize}

\begin{figure*}[htbp]
\centering 
\includegraphics[width=0.97\columnwidth]{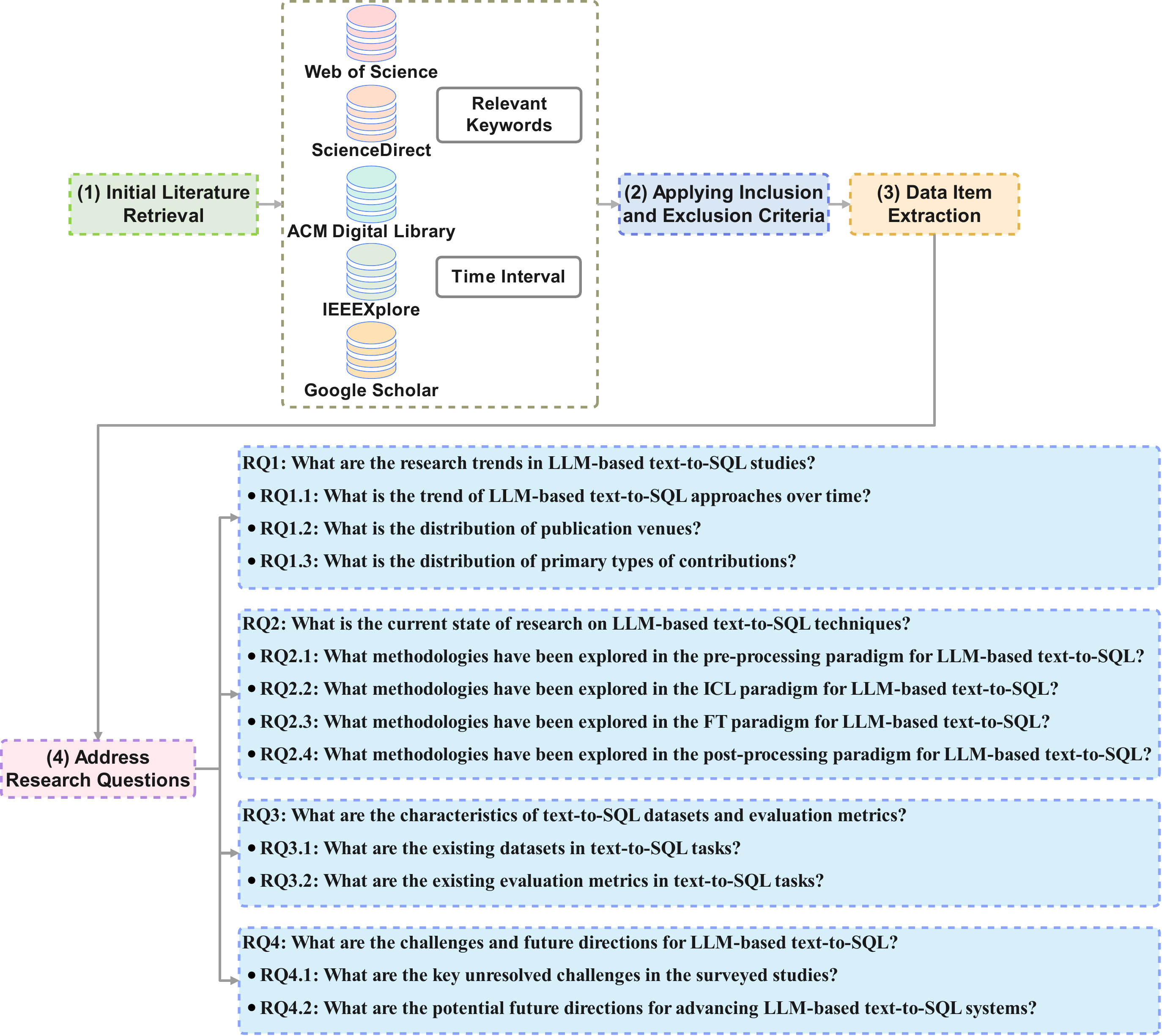}
\caption{The entire process of our systematic survey, including initial literature retrieval, applying inclusion and exclusion criteria, and data item extraction. These steps are used to address the given research questions in this survey.}
\label{fig2}
\end{figure*}

\subsection{Bibliography Retrieval and Selection}
Following the guidelines in \cite{petersen2015guidelines}, we carried out a systematic bibliography retrieval procedure to gather relevant articles. As illustrated in Fig. \ref{fig2}, the entire process includes initial literature retrieval, applying inclusion and exclusion criteria, and data item extraction. These steps contribute to addressing the above research questions. The details of each stage are outlined below.

\subsubsection{Initial Literature Retrieval}
The first step involves retrieving existing articles as comprehensively as possible. To achieve this, the following key aspects should be considered:
\begin{itemize}
    \item \textbf{Retrieval Databases.} In this study, we manually chose several popular and widely adopted digital libraries as our retrieval databases, including Web of Science, ScienceDirect, ACM digital library, IEEEXplore, and Google Scholar.
    \item \textbf{Keyword Retrieval.} To the best of our knowledge, ``\textit{Text-to-SQL}'' and ``\textit{Text2SQL}'' are the most precise terms commonly employed in the literature. Moreover, some researchers also utilized ``\textit{Natural Language-to-SQL}'', ``\textit{NL2SQL}'', and ``\textit{SQL Generation}''. As a result, these terms were selected as retrieval keywords to ensure a broader selection of relevant papers.
    \item \textbf{Time Interval.} In the existing articles, the first study related to LLM-based text-to-SQL approaches was published by \cite{rajkumar2022evaluating} in April 2022. To systematically summarize the progress in LLM-based SQL generation methods, we established the retrieval time interval from April 2022 to October 2024. Additionally, datasets and evaluation metrics regarding text-to-SQL were selected from 2017 to October 2024 to provide a broader introduction for researchers.
\end{itemize}

\subsubsection{Applying Inclusion and Exclusion Criteria}
Given that a large volume of papers with different levels of quality or unrelated topics can be obtained from the first step, we developed the following standards to identify suitable articles and eliminate those that do not pertain to the topic.

\begin{itemize}
    \item The research should be written in English.
    \item The research should focus on innovative LLM-based text-to-SQL methodologies rather than non-LLM approaches. New datasets and evaluation metrics should be related to text-to-SQL tasks.
    \item The time interval regarding LLM-based text-to-SQL methods should be from April 2022 to October 2024, while datasets and evaluation metrics regarding text-to-SQL should be selected from 2017 to October 2024.
    \item The studies should have a minimum length of four pages.
    \item Books, chapters, technical reports, tutorials, and earlier surveys are excluded.
    \item Datasets and evaluation metrics which are not open to the public are not involved.
    \item Conference papers with extended journal versions are discarded, retaining only the journal versions.

\end{itemize}

In this step, we manually examined the title and abstract of each paper to determine its relevance to the research topic of our survey. Finally, we gathered 122 relevant articles that completely satisfied the aforementioned criteria.

\subsubsection{Data Item Extraction}
To address the above research questions, we meticulously reviewed the 122 articles and collected data items listed in Table \ref{data}. Our data item collection primarily centered on the following aspects: publication trend, novel LLM-based approaches, datasets and evaluation metrics, challenges and future directions.

\begin{table*}[htbp] 
\caption{Extracted data items for different research questions.}
\label{data}
  \begin{center}
  \begin{tabular}{c|c|c}
  \toprule
  \scalebox{0.8}{\textbf{RQ}} & \scalebox{0.8}{\textbf{Extracted Data Items}} & \scalebox{0.8}{\textbf{Description}} \\
  \midrule
  \scalebox{0.8}{RQ1} & \scalebox{0.8}{Publication Date/Venue} & \scalebox{0.8}{Basic information of each surveyed article.} \\ 
  \scalebox{0.8}{RQ1} & \scalebox{0.8}{Main Types of Contribution} & \scalebox{0.8}{Main category of contribution in each surveyed article.} \\ \midrule
  \scalebox{0.8}{RQ2} & \scalebox{0.8}{Methodologies for Pre-Processing} & \scalebox{0.8}{Classification for pre-processing methodologies in each surveyed article.} \\
  \scalebox{0.8}{RQ2} & \scalebox{0.8}{Methodologies for ICL} & \scalebox{0.8}{Classification for ICL-based methodologies in each surveyed article.} \\
  \scalebox{0.8}{RQ2} & \scalebox{0.8}{Methodologies for FT} & \scalebox{0.8}{Classification for FT-based methodologies in each surveyed article.} \\
  \scalebox{0.8}{RQ2} & \scalebox{0.8}{Methodologies for Post-Processing} & \scalebox{0.8}{Classification for post-processing methodologies in each surveyed article.} \\ \midrule
  \scalebox{0.8}{RQ3} & \scalebox{0.8}{Datasets} & \scalebox{0.8}{Released datasets for text-to-SQL tasks in each surveyed article.} \\
  \scalebox{0.8}{RQ3} & \scalebox{0.8}{Evaluation Metrics} & \scalebox{0.8}{Proposed evaluation metrics for text-to-SQL tasks in each surveyed article.} \\ \midrule
  \scalebox{0.8}{RQ4} & \scalebox{0.8}{Challenges} & \scalebox{0.8}{Challenges for LLM-based text-to-SQL.} \\
  \scalebox{0.8}{RQ4} & \scalebox{0.8}{Future Directions} & \scalebox{0.8}{Future directions for LLM-based text-to-SQL.} \\
  \bottomrule
  \end{tabular}
  \end{center}
\end{table*}

\section{RQ1: What are the research trends in LLM-based text-to-SQL studies?} \label{RQ1}
In this research question, we examined the fundamental information from the surveyed studies, focusing on publication dates, venues, and primary types of contributions.

\begin{figure*}[htbp]
\centering
\subfigure[Number of articles every two months.]{
		\includegraphics[scale=0.35]{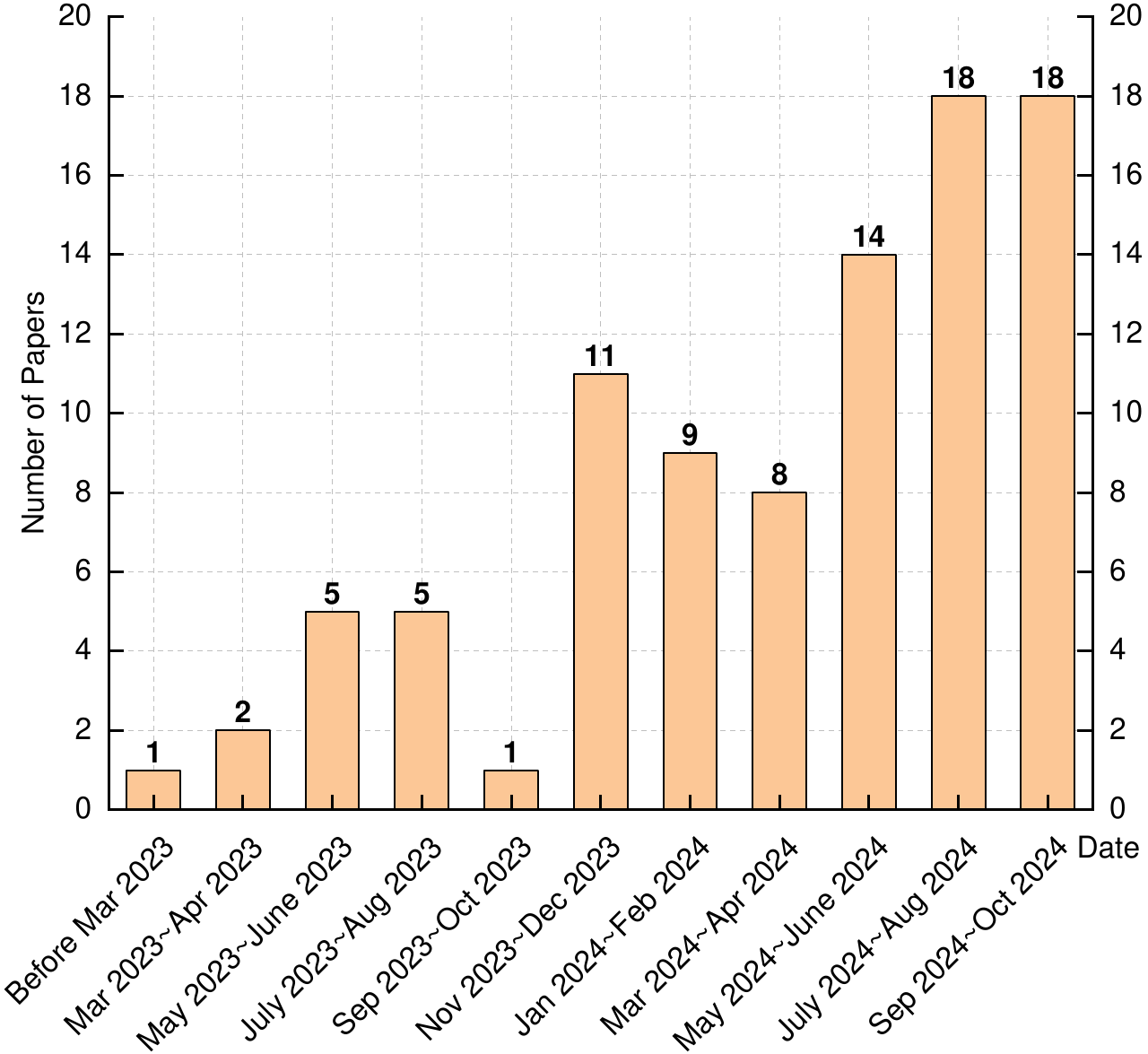}}
\subfigure[Cumulative number of articles every two months.]{
		\includegraphics[scale=0.35]{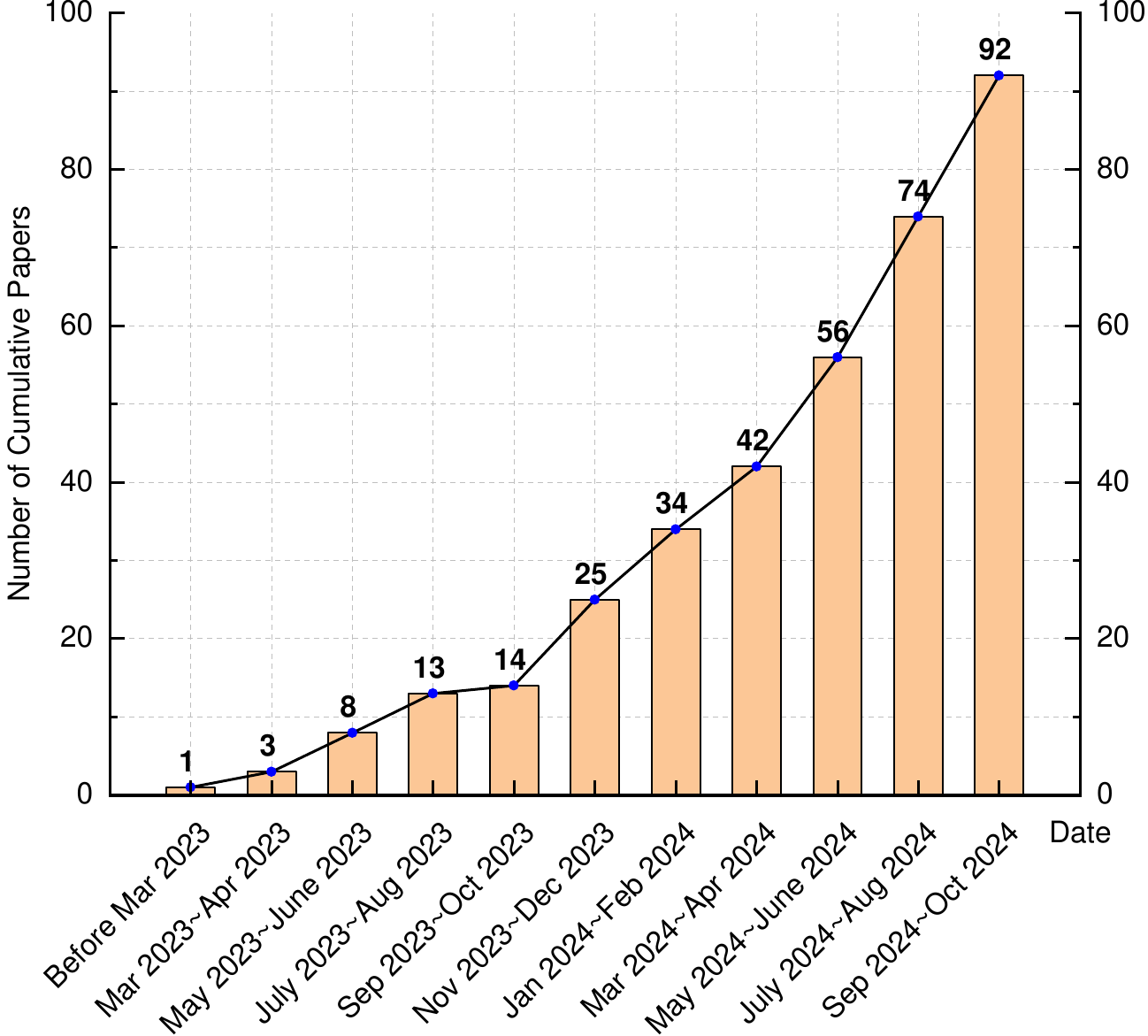}}
\caption{Publication trends of the surveyed LLM-based text-to-SQL approaches from April 2022 to October 2024.}
\label{fig3}
\end{figure*}

\subsection{RQ1.1: What is the trend of LLM-based text-to-SQL approaches over time?}
Fig. \ref{fig3}(a) displays the publication count for LLM-based text-to-SQL approaches every two months. Specifically, the earliest identified study was released in April 2022, marking the beginning of steady growth in this field. Moreover, the number of publications peaked from July 2024 to October 2024, accounting for approximately 39\% of the surveyed LLM-based methods. Fig. \ref{fig3}(b) presents the cumulative publication counts based on Fig. \ref{fig3}(a). Notably, the trend line reveals a marked increase in slope from September 2023 to October 2024, underscoring the growing interest in developing innovative LLM-based text-to-SQL methodologies.

\begin{table*}[htbp] 
\caption{The distribution of 122 surveyed papers in different publication venues.}
\label{venue}
  \begin{center}
  \begin{tabular}{c|c|c}
  \toprule
  \scalebox{0.8}{\textbf{Venue}} & \scalebox{0.8}{\textbf{Full Name}} & \scalebox{0.8}{\textbf{\# Studies}} \\
  \midrule
  \scalebox{0.8}{EMNLP} & \scalebox{0.8}{Conference on Empirical Methods in Natural Language Processing} & \scalebox{0.8}{20}  \\ 
  \scalebox{0.8}{ACL} & \scalebox{0.8}{Annual Meeting of the Association for Computational Linguistics} & \scalebox{0.8}{16}  \\ 
  \scalebox{0.8}{NeurIPS} & \scalebox{0.8}{Conference on Neural Information Processing Systems} & \scalebox{0.8}{7}  \\
  \scalebox{0.8}{NAACL} & \scalebox{0.8}{Annual Conference of the North American Chapter of the Association for Computational Linguistics} & \scalebox{0.8}{4}  \\ 
  \scalebox{0.8}{VLDB} & \scalebox{0.8}{International Conference on Very Large Data Bases} & \scalebox{0.8}{3}  \\ 
  \scalebox{0.8}{ICML} & \scalebox{0.8}{International Conference on Machine Learning} & \scalebox{0.8}{3}  \\
  \scalebox{0.8}{ICLR} & \scalebox{0.8}{International Conference on Learning Representations} & \scalebox{0.8}{3}  \\
  \scalebox{0.8}{SIGMOD} & \scalebox{0.8}{ACM SIGMOD International Conference on Management of Data} & \scalebox{0.8}{2}  \\ 
  \scalebox{0.8}{ICDE} & \scalebox{0.8}{IEEE International Conference on Data Engineering} & \scalebox{0.8}{2}  \\ 
  \scalebox{0.8}{EACL} & \scalebox{0.8}{Conference of the European Chapter of the Association for Computational Linguistics} & \scalebox{0.8}{1}  \\ 
  \scalebox{0.8}{PRICAI} & \scalebox{0.8}{Pacific Rim International Conference on Artificial Intelligence} & \scalebox{0.8}{1}  \\    
  \scalebox{0.8}{ICONIP} & \scalebox{0.8}{International Conference on Neural Information Processing} & \scalebox{0.8}{1}  \\ 
  \scalebox{0.8}{COLING} & \scalebox{0.8}{International Conference on Computational Linguistics} & \scalebox{0.8}{1}  \\ 
  \scalebox{0.8}{ECAI} & \scalebox{0.8}{European Conference on Artificial Intelligence} & \scalebox{0.8}{1}  \\ 
  \scalebox{0.8}{ICSE} & \scalebox{0.8}{International Conference on Software Engineering} & \scalebox{0.8}{1}  \\ 
  \scalebox{0.8}{AAAI} & \scalebox{0.8}{Proceedings of the AAAI conference on artificial intelligence} & \scalebox{0.8}{1}  \\
  \scalebox{0.8}{WWW} & \scalebox{0.8}{International World Wide Web Conference} & \scalebox{0.8}{1}  \\
  \scalebox{0.8}{UIST} & \scalebox{0.8}{Annual ACM Symposium on User Interface Software and Technology} & \scalebox{0.8}{1}  \\
  \scalebox{0.8}{DEXA} & \scalebox{0.8}{International Conference on Database And Expert System Applications} & \scalebox{0.8}{1}  \\
  \midrule
  \scalebox{0.8}{DSE} & \scalebox{0.8}{Data Science and Engineering} & \scalebox{0.8}{1} \\ 
  \scalebox{0.8}{TCDS} & \scalebox{0.8}{IEEE Transactions on Cognitive and Developmental Systems} & \scalebox{0.8}{1} \\
  \scalebox{0.8}{IEEE ACCESS} & \scalebox{0.8}{-} & \scalebox{0.8}{1} \\
  \midrule
  \scalebox{0.8}{Others} & \scalebox{0.8}{Arxiv} & \scalebox{0.8}{49}  \\ 
  \bottomrule
  \end{tabular}
  \end{center}
\end{table*}

\subsection{RQ1.2: What is the distribution of publication venues?}
In this survey, a systematic review was conducted on 122 surveyed papers, encompassing different approaches, datasets, and evaluation metrics from multiple publication venues. It comprises 70 articles from conferences and symposiums, 3 articles from journals, and 49 articles from Arxiv that have not been further published in conferences or journals. Table \ref{venue} illustrates the distribution of the collected papers across these venues. In particular, publication venues were ranked in descending order based on the number of the surveyed papers. From Table \ref{venue}, we can draw the following observations:

(1) The majority of the surveyed articles were published in conference proceedings instead of in journals. Specifically, conference papers account for 57\% of the total, while only 2\% of the papers were published in journals with peer-review. This trend may be partially attributed to the more extensive and rigorous review process associated with journals, encouraging researchers to favor conference publications for their timeliness. Moreover, a significant number of text-to-SQL studies have been published in Arxiv rather than in conferences or journals. The reason is that the rigorous review process of conferences and journals often results in lengthy publication timelines. In contrast, Arxiv enables researchers to share their findings within days, making it an appealing option for those aiming to quickly introduce their recent research and gather early feedback from the community. 

(2) A substantial number of publication venues (22 in total) are relevant for the surveyed articles, involving 19 conferences and 3 journals. It is evident that \textit{EMNLP} and \textit{ACL} are the most prominent venues with the highest number of studies, since the topic of text-to-SQL is more closely related to NLP and artificial intelligence. 

\begin{table*}[htbp] 
\caption{The description and distribution of main contributions in the surveyed text-to-SQL articles. Note that the percentage for each category is the number of papers in that category compared to the 122 surveyed papers in total.}
\label{contribution}
  \begin{center}
  \begin{tabular}{c|l|c}
  \toprule
  \scalebox{0.8}{\textbf{Main contribution}} & \scalebox{0.8}{\textbf{Description}} & \scalebox{0.8}{\textbf{\# (\%)}} \\
  \midrule
  \scalebox{0.8}{LLM-based Text-to-SQL Approaches} & \scalebox{0.8}{\makecell[l]{The article presented a novel LLM-based SQL generation technique. It detailed \\ a comprehensive methodology and included extensive experiments to assess \\ the efficacy of this approach.}} & \scalebox{0.8}{92 (75\%)} \\  
  \scalebox{0.8}{Newly-Released Datasets} & \scalebox{0.8}{\makecell[l]{The paper introduced a new dataset for text-to-SQL tasks, which was open to \\ the public.}} & \scalebox{0.8}{26 (21\%)} \\
  \scalebox{0.8}{Newly-Introduced Evaluation Metrics} & \scalebox{0.8}{\makecell[l]{The paper introduced a new metric to assess the model performance for SQL \\ generation, which was open to the public.}} & \scalebox{0.8}{7 (6\%)} \\ 
  \bottomrule
  \end{tabular}
  \end{center}
\end{table*}

\subsection{RQ1.3: What is the distribution of primary types of contributions?}
\begin{figure*}[htbp]
\centering 
\includegraphics[width=0.97\columnwidth]{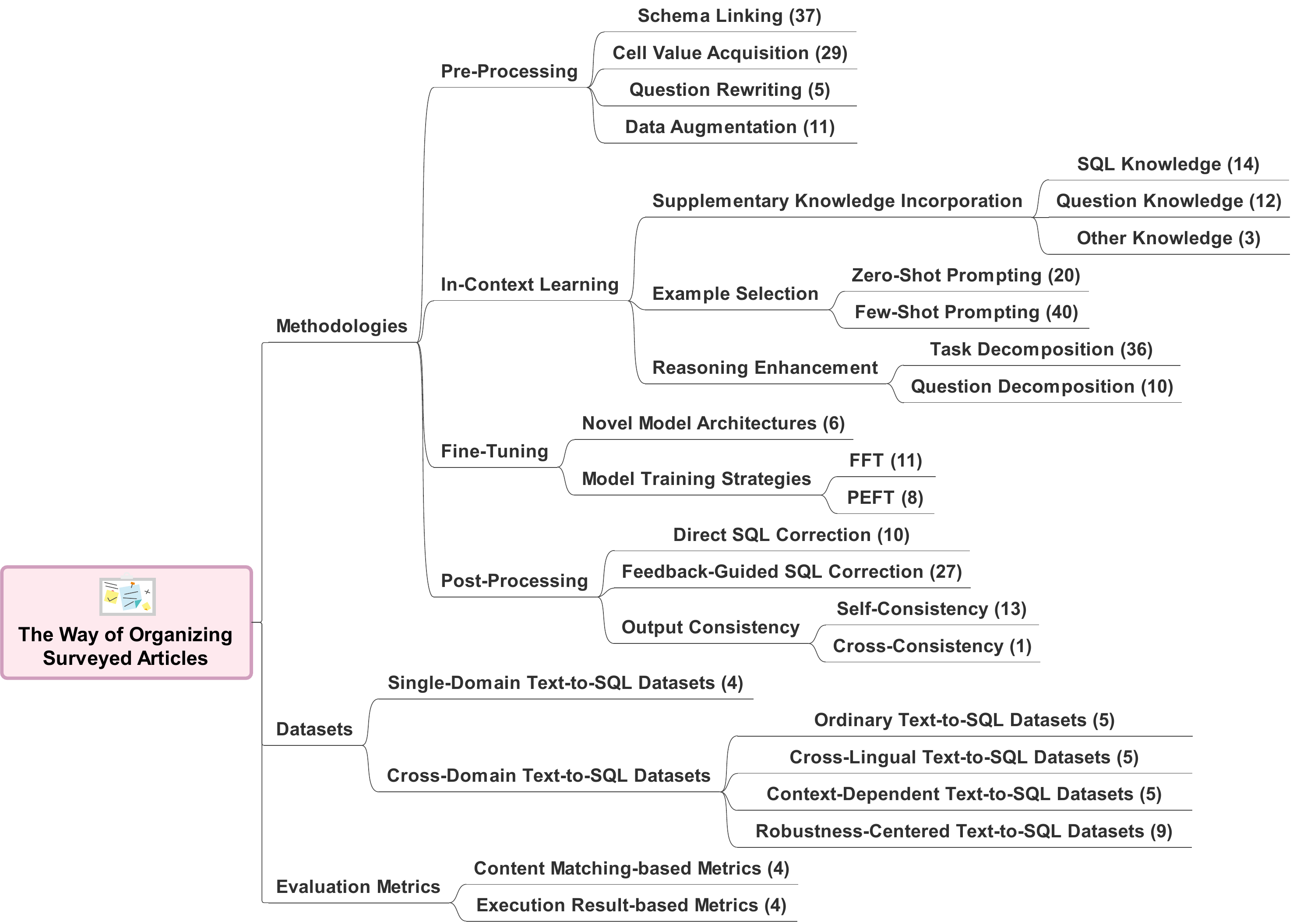}
\caption{Overview of the way of organizing surveyed articles. Note that the number in parentheses represents the number of the surveyed articles in each category.}
\label{fig4}
\end{figure*}

We manually determined the primary contributions of each surveyed study and classified them accordingly. Table \ref{contribution} presents the definitions of these contributions along with their distribution, including the count and percentage of the surveyed papers in each category. Specifically, the contributions of the reviewed articles can be grouped into three types: \textit{Methodologies}, \textit{Datasets}, and \textit{Evaluation Metrics}. In particular, there are 92 articles proposing LLM-based text-to-SQL approaches, 26 articles introducing new datasets for text-to-SQL tasks, and 7 articles presenting novel evaluation metrics for assessing SQL quality. Fig. \ref{fig4} illustrates the organization of the surveyed articles, and Sections \ref{RQ2}, \ref{RQ3}, and \ref{RQ4} provide a detailed introduction based on this taxonomy.

\begin{tcolorbox}[colback=black!5!white, colframe=black, 
                  title=\textcolor{black}{\textbf{Answer to RQ1}}, 
                  coltitle=white, colbacktitle=pink, 
                  boxrule=1pt, width=\textwidth]
(1) The employment of LLM-based techniques for SQL generation has experienced a notable upward trend from September 2023 to October 2024, showcasing considerable interest in designing LLM-based SQL generation systems.

(2) The majority of text-to-SQL articles have been published in conferences rather than in journals, especially in conferences related to NLP and artificial intelligence.

(3) The studies placed greater emphasis on developing innovative methodologies over introducing new datasets and evaluation metrics, indicating a strong research focus on improving the effectiveness and efficiency of text-to-SQL conversion.
\end{tcolorbox}

\section{RQ2: What is the current state of research on LLM-based text-to-SQL techniques?} \label{RQ2}
\begin{figure*}[htbp]
\centering 
\includegraphics[width=1.0\columnwidth]{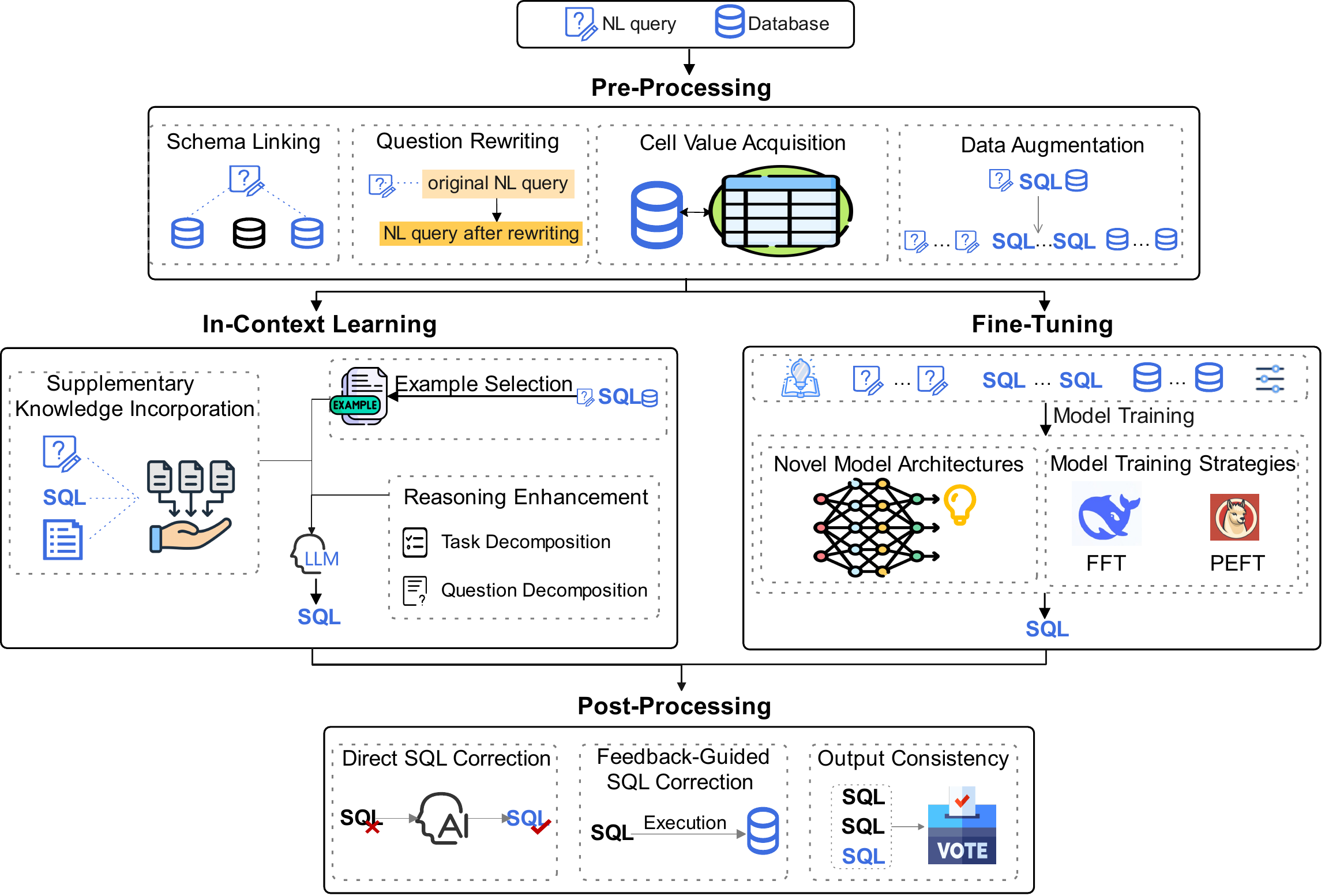}
\caption{Illustration of a typical LLM-based text-to-SQL workflow, involving pre-processing, in-context learning, fine-tuning and post-processing paradigms.}
\label{fig5}
\end{figure*}

In this research question, an in-depth introduction of the current LLM-based text-to-SQL solutions is provided, focusing on key stages of a typical SQL generation workflow illustrated in Fig. \ref{fig5}. This process encompasses pre-processing, in-context learning, fine-tuning, and post-processing, each of which plays an essential role in achieving accurate SQL generation.

\subsection{RQ2.1: What methodologies have been explored in the pre-processing paradigm for LLM-based text-to-SQL?}
The pre-processing paradigm is crucial for preparing and refining input data to ensure accurate SQL generation. It can be classified into \textit{Schema Linking}, \textit{Cell Value Acquisition}, \textit{Question Rewriting}, and \textit{Data Augmentation}, as shown in Table \ref{pre}.

\textbf{Schema Linking.} The complexity of databases and data structures presents a notable challenge in SQL generation, particularly in accurately identifying relevant tables and columns. Schema linking, which maps words or phrases from user queries to database schema elements (e.g., tables, columns, and relationships), has been a focal point in text-to-SQL systems. Numerous schema linking methods have been developed in the literature \cite{pourreza2024din, dong2023c3, wang2024mac, lee2024mcs, ren2024purple}. For instance, DIN-SQL \cite{pourreza2024din} utilized the Chain-of-Thought (CoT) reasoning technique to create schema-linking prompts. This step-by-step reasoning process effectively links NL keywords to relevant database tables and columns. C3 \cite{dong2023c3} used zero-shot prompting to rank tables and columns related to NL questions for schema linking. MAC-SQL \cite{wang2024mac} employed a Selector to decompose a large database into smaller sub-databases for selecting relevant tables and columns, while MCS-SQL \cite{lee2024mcs} enhanced schema linking by leveraging multiple prompts, allowing for a more comprehensive exploration of relevant tables and columns. PURPLE \cite{ren2024purple} adopted trained classifiers to link questions to database schemas and reduced irrelevant information through pattern pruning, streamlining schema linking for more efficient SQL generation. 

\textbf{Cell Value Acquisition.} Different from schema linking focusing on identifying pertinent tables and columns, cell value acquisition emphasizes the efficient extraction of relevant cell values from the database. It provides context-rich information directly to the LLMs, which is essential when NL queries lack sufficient detail for accurate SQL generation. Several approaches have been developed to enhance cell value acquisition for text-to-SQL tasks \cite{li2024pet, sun2023sql, xu2024tcsr, mao2024enhancing, li2024sea, li2024codes}. For instance, PET-SQL \cite{li2024pet} embedded a random selection of database values directly into the prompt, offering the model additional context to improve SQL accuracy. SQL-PaLM \cite{sun2023sql} applied the longest contiguous matching sub-sequence approach \cite{cormen2022introduction} to calculate keyword similarity between NL query and database values. Irrelevant matches were filtered using similarity thresholds, which improved the relevance of retrieved cell values. TCSR-SQL \cite{xu2024tcsr} aligned keywords in NL queries with stored database values through an encoding knowledge table, ensuring that the generated SQL is based on the most relevant data. DART-SQL \cite{mao2024enhancing} calculated cosine similarity between database rows with cell values and questions using GloVe embeddings \cite{pennington2014glove}. It then selected the most relevant rows and a subset of random rows to provide the model with additional context. SEA-SQL \cite{li2024sea} and CodeS \cite{li2024codes} constructed a BM25 index from database content, retrieving the top-$k$ values from each relevant column to improve semantic representation in SQL generation.

\textbf{Question Rewriting.} Due to the inherent semantic ambiguity in NL questions and the misalignment between NL queries and database content, several studies \cite{guo2023retrieval, sui2023reboost, mao2024enhancing, yan2024intelliexplain, caferolu2024esql} have focused on question rewriting to facilitate accurate SQL generation. Specifically, \cite{guo2023retrieval} simplified NL queries by prompting LLMs with manual instructions, enabling the questions to be clear and easy to understand. ReBoostSQL \cite{sui2023reboost} identified ambiguous terms in questions and converted them into explicitly defined expressions to improve clarity. DART-SQL \cite{mao2024enhancing} used NL questions and relevant database information as inputs, guiding LLMs in rewriting questions clearly, concisely, and well-aligned with database specifications. Similarly, E-SQL \cite{caferolu2024esql} directed LLMs to refine the original NL queries by incorporating relevant database schema information, thereby making the queries clearer, more comprehensible, and free of unnecessary details. IntelliExplain \cite{yan2024intelliexplain} leveraged LLMs to rewrite NL queries in a more straightforward manner, allowing users to verify whether the system has accurately interpreted the question and to make adjustments based on the system's understanding. 

\textbf{Data Augmentation.} The effectiveness of LLMs is heavily influenced by the quantity and quality of their training data \cite{deng2022recent}, underscoring the significance of data augmentation in optimizing performance for text-to-SQL tasks. Recently, novel semi-automatic and fully automated data augmentation approaches have emerged, leveraging LLMs' generative capabilities to create synthetic datasets for training other models or enhancing the LLMs themselves \cite{sun2023sql, li2024codes, saparina2024improving, xu2023symbol, zhuang2024structlm, wu2024datagptsql}. For instance, SQL-PaLM \cite{sun2023sql} generated new SQL queries different from the ground truths while preserving the same database schema and NL questions. Similarly, \cite{saparina2024improving} utilized LLMs to produce more realistic and diverse questions while keeping the corresponding SQL and database schema consistent. CodeS \cite{li2024codes} introduced a bi-directional data augmentation approach to automatically generate a diverse set of (NL, SQL) pairs. Symbol-LLM \cite{xu2023symbol} implemented a two-stage data augmentation strategy with injection and infusion phases, enhancing the model's capacity for tackling NL-centric tasks like SQL generation. StructLM \cite{zhuang2024structlm} applied multi-task learning to train on a range of structured knowledge tasks, enriching the model’s understanding of structural patterns essential for SQL generation. DATAGPT-SQL \cite{wu2024datagptsql} introduced cross-database and inner-database augmentation techniques, expanding datasets to enhance the model's capability to identify the correct tables and columns in database schemas.

\begin{table*}[htbp] 
\caption{A summary of LLM-based methods employed in the pre-processing paradigm for text-to-SQL tasks. These methods are categorized based on their specific implementation designs. Note that SL-Schema Linking, CVA-Cell Value Acquisition, QR-Question Rewriting, DA-Data Augmentation.}
\label{pre}
  \begin{center}
  \begin{tabular}{c|c|l}  
  \toprule
  \scalebox{0.8}{\textbf{Category}} & \scalebox{0.8}{\textbf{\# Studies}} & \scalebox{0.8}{\textbf{References}} \\
  \midrule
  \scalebox{0.8}{SL} & \scalebox{0.8}{37} & \makecell[l]{\scalebox{0.8}{\cite{li2024codes}}, \scalebox{0.8}{\cite{pourreza2024dts}}, \scalebox{0.8}{\cite{zhang2024sqlfuse}}, \scalebox{0.8}{\cite{zhang2024finsql}}, \scalebox{0.8}{\cite{gorti2024msc}}, \scalebox{0.8}{\cite{dong2023c3}}, \scalebox{0.8}{\cite{pourreza2024din}}, \scalebox{0.8}{\cite{kothyari2023crush4sql}}, \scalebox{0.8}{\cite{guo2023prompting}}, \scalebox{0.8}{\cite{sun2023sql}}, \scalebox{0.8}{\cite{sui2023reboost}}, \scalebox{0.8}{\cite{liu2023divide}}, \scalebox{0.8}{\cite{li2024pet}}, \scalebox{0.8}{\cite{wang2024mac}}, \scalebox{0.8}{\cite{zhang2023act}}, \scalebox{0.8}{\cite{xie2024decomposition}}, \scalebox{0.8}{\cite{ren2024purple}}, \scalebox{0.8}{\cite{wang2024dac}}, \\
  \scalebox{0.8}{\cite{xie2024mag}}, \scalebox{0.8}{\cite{volvovsky2024dfin}}, \scalebox{0.8}{\cite{shen2024improving}},  \scalebox{0.8}{\cite{xu2024tcsr}}, \scalebox{0.8}{\cite{cen2024sqlfix}}, \scalebox{0.8}{\cite{talaei2024chess}}, \scalebox{0.8}{\cite{qu2024before}}, \scalebox{0.8}{\cite{lee2024mcs}}, \scalebox{0.8}{\cite{li2024using}}, \scalebox{0.8}{\cite{zhang2024structure}}, \scalebox{0.8}{\cite{wang2024tool}}, \scalebox{0.8}{\cite{tai2023exploring}}, \scalebox{0.8}{\cite{chen2024teaching}}, \scalebox{0.8}{\cite{mao2024enhancing}}, \scalebox{0.8}{\cite{tan2024enhancing}}, \scalebox{0.8}{\cite{shen2024selectsql}}, \scalebox{0.8}{\cite{luo2024ptdsql}}, \scalebox{0.8}{\cite{caferolu2024esql}}, \\ \scalebox{0.8}{\cite{sun2024linking}}} \\ 
  
  \scalebox{0.8}{CVA} & \scalebox{0.8}{29} & \makecell[l]{\scalebox{0.8}{\cite{li2024codes}}, \scalebox{0.8}{\cite{pourreza2024dts}}, \scalebox{0.8}{\cite{yang2024synthesizing}}, \scalebox{0.8}{\cite{li2024sea}}, \scalebox{0.8}{\cite{zhang2024sqlfuse}}, \scalebox{0.8}{\cite{gorti2024msc}}, \scalebox{0.8}{\cite{zhong2024learning}}, \scalebox{0.8}{\cite{sun2023sql}}, \scalebox{0.8}{\cite{arora2023adapt}}, \scalebox{0.8}{\cite{chang2023selective}}, \scalebox{0.8}{\cite{sun2023sqlprompt}}, \scalebox{0.8}{\cite{chang2023prompt}}, \scalebox{0.8}{\cite{li2024pet}}, \scalebox{0.8}{\cite{wang2024mac}}, \scalebox{0.8}{\cite{zhang2023act}}, \scalebox{0.8}{\cite{wang2024dac}}, \scalebox{0.8}{\cite{xie2024mag}}, \scalebox{0.8}{\cite{shen2024improving}}, \\ \scalebox{0.8}{\cite{xu2024tcsr}}, \scalebox{0.8}{\cite{talaei2024chess}}, \scalebox{0.8}{\cite{qu2024before}}, \scalebox{0.8}{\cite{thorpe2024dubo}}, \scalebox{0.8}{\cite{li2024using}}, \scalebox{0.8}{\cite{chen2024teaching}}, \scalebox{0.8}{\cite{gu2024middleware}}, \scalebox{0.8}{\cite{mao2024enhancing}}, \scalebox{0.8}{\cite{wang2024tool}}, \scalebox{0.8}{\cite{caferolu2024esql}}, \scalebox{0.8}{\cite{pourreza2024chasesql}}} \\  
  
  \scalebox{0.8}{QR} & \scalebox{0.8}{5} & \scalebox{0.8}{\cite{sui2023reboost}}, \scalebox{0.8}{\cite{guo2023retrieval}}, \scalebox{0.8}{\cite{yan2024intelliexplain}}, \scalebox{0.8}{\cite{mao2024enhancing}}, \scalebox{0.8}{\cite{caferolu2024esql}} \\  
  \scalebox{0.8}{DA} & \scalebox{0.8}{11} & \scalebox{0.8}{\cite{sun2023sql}}, \scalebox{0.8}{\cite{saparina2024improving}}, \scalebox{0.8}{\cite{pourreza2024chasesql}}, \scalebox{0.8}{\cite{li2024codes}}, \scalebox{0.8}{\cite{pourreza2024sql}}, \scalebox{0.8}{\cite{yang2024synthesizing}}, \scalebox{0.8}{\cite{zhang2024finsql}}, \scalebox{0.8}{\cite{xu2023symbol}}, \scalebox{0.8}{\cite{zhuang2024structlm}}, \scalebox{0.8}{\cite{kobayashi2024yoro}}, \scalebox{0.8}{\cite{wu2024datagptsql}} \\  
  \bottomrule
  \end{tabular}
  \end{center}
\end{table*}

\subsection{RQ2.2: What methodologies have been explored in the ICL paradigm for LLM-based text-to-SQL?}
In the literature, in-context learning (ICL)-based approaches primarily focus on prompt engineering techniques, which reduce computational costs by eliminating the need for additional training data. Meanwhile, these approaches allow for adaptability across different scenarios through modifications to manually crafted prompts. However, different prompt designs can significantly impact the performance of LLMs in text-to-SQL tasks \cite{yang2024harnessing}. Consequently, developing text-to-SQL methodologies within the ICL framework is essential for achieving significant advancements. As shown in Table \ref{ICL}, various meticulously crafted prompt designs have been proposed for text-to-SQL tasks, which can be generally classified into the following categories.

\textbf{Supplementary Knowledge Incorporation.} To enhance LLM's performance, numerous studies have integrated supplementary knowledge into prompting strategies, increasing the accuracy of the generated SQL. This knowledge can generally be divided into the following categories:

(1) \textbf{SQL Knowledge.}
SQL is a standardized programming language for managing relational databases, characterized by its highly structured and rule-based nature. To enhance LLM's understanding of SQL syntax, previous approaches \cite{zhang2024benchmarking, pourreza2024din, wang2024large, nan2023enhancing} have introduced specific rules or templates about SQL queries, collectively referred to as SQL knowledge. For instance, \cite{zhang2024benchmarking} utilized various SQL templates to facilitate the production of multiple candidate SQL queries, thereby increasing the likelihood of generating effective SQL. DIN-SQL \cite{pourreza2024din} and MLPrompt \cite{wang2024large} analyzed the erroneous SQL queries produced by LLMs and established a collection of guidelines to improve SQL generation. \cite{nan2023enhancing} divided the given annotated examples based on SQL difficulty, subsequently applying $k$-Means \cite{he2024application} in selecting examples for each category.

(2) \textbf{Question Knowledge.}
Several studies \cite{thorpe2024dubo, gao2023text, li2024pet, guo2023retrieval} have explored problem mining to extract useful knowledge from NL questions. This knowledge was then integrated with few-shot learning by embedding similar NL queries into prompts to enhance LLM reasoning. For instance, Dubo-SQL \cite{thorpe2024dubo} and DAIL-SQL \cite{gao2023text} used a text-embedding model to calculate vector embeddings for NL questions in the training set, followed by the selection of pertinent examples based on cosine similarity. PET-SQL \cite{li2024pet} employed the structural skeleton of NL queries to obtain similar (NL, SQL) pairs, aiming to enhance adaptability across diverse NL queries by emphasizing structure over specific details. Similarly, \cite{guo2023retrieval} removed schema-related tokens from questions, highlighting the core structure and intent of NL queries. The extracted question skeletons were subsequently used to retrieve more relevant examples.

(3) \textbf{Other Knowledge.}
In text-to-SQL tasks, additional knowledge can also be integrated into prompts, such as database knowledge \cite{ma2024enhancing}, symbolic memory knowledge \cite{hu2023chatdb} and task-related knowledge \cite{li2024can}. Specifically, KE \cite{ma2024enhancing} proposed a method for injecting database knowledge, incorporating schema information into LLMs through targeted training objectives. ChatDB \cite{hu2023chatdb} employed symbolic memory knowledge to support LLMs in multi-step reasoning by decomposing complicated database operations into sequential execution. This approach stored the SQL execution’s results in memory, using this stored information to determine subsequent operations. Moreover, specific task-related knowledge such as the calculation for ``price per unit of product'' in customer purchase scenario was integrated into prompts \cite{li2024can}. This incorporation allows LLMs to produce more precise domain-specific SQL queries.

\textbf{Example Selection.} Incorporating relevant (NL, SQL) examples is crucial for achieving satisfactory performance in LLM-based text-to-SQL tasks. Similar to humans reflecting on past experiences to make accurate decisions, LLMs can benefit from pertinent examples to improve the text-to-SQL translation. In the literature, example selection approaches are typically divided into \textit{Zero-Shot} and \textit{Few-Shot} settings.

(1) \textbf{Zero-Shot Prompting.} Several methods \cite{chang2023prompt, dong2023c3, liu2023comprehensive, liu2023divide, sui2023reboost} have investigated zero-shot strategies, which avoid including specific text-to-SQL examples in prompts. Instead, these approaches centered on other aspects, such as understanding database schemas or decomposing text-to-SQL into multiple stages. By leveraging LLM's internal knowledge and problem-solving capabilities, zero-shot methods enable SQL generation without relying on explicit examples. This setting is particularly advantageous when domain-specific examples are unavailable or when evaluating a model's capacity to apply general reasoning skills across new domains.

(2) \textbf{Few-Shot Prompting.} Few-shot prompting has become central in optimizing LLM-based SQL generation performance, focusing on selecting a limited number of examples that effectively guide the model. Various approaches  \cite{ni2023lever, li2024pet, gao2023text, shen2024improving, thorpe2024dubo, tai2023exploring, zhang2023act, arora2023adapt, chang2023selective} have been developed to retrieve these few-shot examples. 
\begin{itemize}
    \item \textit{Random Example Selection} typically serves as a baseline method by selecting examples without specific criteria. Studies like LEVER \cite{ni2023lever} utilized random selection to demonstrate the baseline performance of few-shot prompting.
    \item \textit{Similarity-based Example Selection} retrieves examples that are semantically similar to the given query. For instance, PET-SQL \cite{li2024pet} and DAIL-SQL \cite{gao2023text} employed a sentence embedding model for semantic similarity, while \cite{shen2024improving} utilized Abstract Syntax Tree (AST)-based ranking \cite{xiang2024sql} to ensure that the selected examples were contextually relevant. 
    \item \textit{Difficulty-based Example Selection} chooses examples based on task complexity \cite{tai2023exploring}, which facilitates model adaptability to tasks with different difficulty levels.
    \item \textit{Hybrid Example Selection} combines static and dynamically selected examples to provide balanced and effective prompts for consistent and adaptive guidance. For instance, ACT-SQL \cite{zhang2023act} used randomly chosen examples from the training set, which remain constant across all test cases to ensure baseline guidance. Additionally, it incorporated dynamic examples selected via \textit{Similarity-based Example Selection}, customizing the prompt to the specific characteristics of each test case for improved context relevance.
    \item \textit{Cross- and In-domain Example Selection} investigates the benefits of incorporating examples from in-domain and out-of-domain sources. For instance, DA-GP \cite{arora2023adapt} employed cross-domain examples to bridge knowledge transfer, while ODIS \cite{chang2023selective} integrated out-of-domain examples with in-domain data for comprehensive task and domain coverage.
\end{itemize}

\textbf{Reasoning Enhancement.} LLMs have shown significant promise in handling symbolic, commonsense, and arithmetic scenarios owing to their remarkable reasoning capabilities \cite{huang2023towards}. Therefore, effective reasoning enhancement strategies can be leveraged to substantially improve SQL accuracy. Notably, recent studies have introduced Decomposition techniques to address the complexity of intricate text-to-SQL tasks. This technique enables LLMs to construct SQL progressively, enhancing reasoning capabilities by breaking down complex tasks or questions. In the literature, it can primarily be divided into \textit{Task Decomposition} and \textit{Question Decomposition}.

(1) \textbf{Task Decomposition.} This involves segmenting the text-to-SQL process into smaller and manageable steps, allowing each sub-task to address particular challenges in the SQL generation workflow. This approach facilitates precise management of critical steps, such as schema linking and SQL formation \cite{pourreza2024din, dong2023c3, xie2024decomposition, wang2024dac}. For instance, DIN-SQL \cite{pourreza2024din} divided the SQL generation process into four modules, namely schema linking, classification and decomposition, SQL generation, and self-correction. C3 \cite{dong2023c3} organized the whole process into three modules: clear prompting, calibration bias prompting, and consistency checks. DEA-SQL \cite{xie2024decomposition} implemented a structured workflow for SQL generation, involving gathering database information, identifying query types, devising solution strategies, generating SQL syntax, conducting initial self-checks, and reviewing past errors to mitigate repetitive mistakes. DAC \cite{wang2024dac} applied task decomposition during the SQL correction phase, splitting the task into entity linking and skeleton parsing stages. By dividing SQL generation into specialized components, task decomposition methods can improve the effectiveness of LLMs in SQL generation.

(2) \textbf{Question Decomposition.} This technique involves breaking user queries into simpler sub-questions, enhancing LLMs' reasoning capabilities in text-to-SQL \cite{zhang2023act, wang2024mac, xie2024mag, pourreza2024chasesql}. For instance, ACT-SQL \cite{zhang2023act} leveraged CoT by automatically breaking down complex NL questions into corresponding SQL components. This ensures that each part of the input is directly linked to its SQL counterpart, contributing to more precise SQL generation. MAC-SQL \cite{wang2024mac} used a Decomposer agent to segment the original user query into smaller sub-questions, facilitating SQL generation for each segment. MAG-SQL \cite{xie2024mag} employed a Targets-Conditions Decomposition method based on the Least-to-Most Prompting approach \cite{zhou2022least}, where each sub-question was generated by incrementally adding conditions to the previous one. Meanwhile, it included a Sub-SQL Refiner to offer refinement at each intermediate SQL generation stage. CHASE-SQL \cite{pourreza2024chasesql} decomposed the NL query into smaller sub-problems, followed by the aggregation of the solutions to these sub-problems. A SQL optimization step was further conducted to eliminate redundant clauses and conditions in the produced SQL.

\begin{table*}[h!]
\centering
\caption{A summary of LLM-based methods employed in the ICL paradigm for text-to-SQL tasks. These methods are categorized based on their specific implementation designs. Note that SKI-Supplementary Knowledge Incorporation, ES-Example Selection, RE-Reasoning Enhancement.}
\label{ICL}
  \begin{center}
  \begin{tabular}{c|c|c|l}
  \toprule
  \scalebox{0.8}{\textbf{Category}} & \scalebox{0.8}{\textbf{Type}} & \scalebox{0.8}{\textbf{\# Studies}} & \scalebox{0.8}{\textbf{References}} \\
  \midrule
  \multirow{3}{*}{\scalebox{0.8}{SKI}} & \scalebox{0.8}{SQL Knowledge} & \scalebox{0.8}{14} &\scalebox{0.8}{\cite{zhang2024benchmarking}}, \scalebox{0.8}{\cite{dong2023c3}}, \scalebox{0.8}{\cite{pourreza2024din}}, \scalebox{0.8}{\cite{gao2023text}}, \scalebox{0.8}{\cite{xie2024decomposition}}, \scalebox{0.8}{\cite{ren2024purple}}, \scalebox{0.8}{\cite{wang2024dac}}, \scalebox{0.8}{\cite{xu2024tcsr}}, \scalebox{0.8}{\cite{wang2024large}}, \scalebox{0.8}{\cite{caferolu2024esql}}, \scalebox{0.8}{\cite{arora2023adapt}}, \scalebox{0.8}{\cite{gu2023few}}, \scalebox{0.8}{\cite{liu2023divide}}, \scalebox{0.8}{\cite{nan2023enhancing}}
  \\  & \scalebox{0.8}{Question Knowledge} & \scalebox{0.8}{12} & \scalebox{0.8}{\cite{thorpe2024dubo}}, \scalebox{0.8}{\cite{li2024pet}}, \scalebox{0.8}{\cite{guo2023retrieval}}, \scalebox{0.8}{\cite{gao2023text}}, \scalebox{0.8}{\cite{nan2023enhancing}}, \scalebox{0.8}{\cite{chang2023selective}}, \scalebox{0.8}{\cite{ren2024purple}},  \scalebox{0.8}{\cite{guo2023prompting}}, \scalebox{0.8}{\cite{xie2024decomposition}}, \scalebox{0.8}{\cite{sui2023reboost}}, \scalebox{0.8}{\cite{arora2023adapt}}, \scalebox{0.8}{\cite{mai2024learning}} \\
   & \scalebox{0.8}{Other Knowledge} & \scalebox{0.8}{3} & \scalebox{0.8}{\cite{ma2024enhancing}}, \scalebox{0.8}{\cite{hu2023chatdb}}, \scalebox{0.8}{\cite{li2024can}} \\
  \midrule
  \multirow{2}{*}{\scalebox{0.8}{ES}} & \scalebox{0.8}{Zero-shot Prompting} & \scalebox{0.8}{20}  & \makecell[l]{\scalebox{0.8}{\cite{dong2023c3}}, \scalebox{0.8}{\cite{guo2023prompting}}, \scalebox{0.8}{\cite{hu2023chatdb}}, \scalebox{0.8}{\cite{sui2023reboost}}, \scalebox{0.8}{\cite{liu2023comprehensive}},  \scalebox{0.8}{\cite{wang2023dbcopilot}}, \scalebox{0.8}{\cite{guo2023retrieval}}, \scalebox{0.8}{\cite{gu2023interleaving}}, \scalebox{0.8}{\cite{fan2024metasql}}, \scalebox{0.8}{\cite{wu2024need}}, \scalebox{0.8}{\cite{cen2024sqlfix}}, \scalebox{0.8}{\cite{askari2024magic}}, \scalebox{0.8}{\cite{yan2024intelliexplain}}, \scalebox{0.8}{\cite{li2024using}}, \\ \scalebox{0.8}{\cite{zhang2024structure}}, \scalebox{0.8}{\cite{volvovsky2024dfin}}, \scalebox{0.8}{\cite{wang2024tool}}, \scalebox{0.8}{\cite{li2024sea}}, \scalebox{0.8}{\cite{wang2024large}}, \scalebox{0.8}{\cite{zheng2024actor}}} \\ 
   & \scalebox{0.8}{Few-shot Prompting} & \scalebox{0.8}{40}  & \makecell[l]{\scalebox{0.8}{\cite{rajkumar2022evaluating}}, \scalebox{0.8}{\cite{pourreza2024din}}, \scalebox{0.8}{\cite{kothyari2023crush4sql}}, \scalebox{0.8}{\cite{sun2023sql}}, \scalebox{0.8}{\cite{liu2023divide}}, \scalebox{0.8}{\cite{arora2023adapt}}, \scalebox{0.8}{\cite{gao2023text}}, \scalebox{0.8}{\cite{nan2023enhancing}}, \scalebox{0.8}{\cite{cheng2022binding}}, \scalebox{0.8}{\cite{tai2023exploring}}, \scalebox{0.8}{\cite{chang2023prompt}}, \scalebox{0.8}{\cite{chang2023selective}}, \scalebox{0.8}{\cite{sun2023sqlprompt}}, \scalebox{0.8}{\cite{ni2023lever}}, \\ \scalebox{0.8}{\cite{li2024pet}}, \scalebox{0.8}{\cite{chen2024teaching}},  \scalebox{0.8}{\cite{gu2024middleware}}, \scalebox{0.8}{\cite{wang2024mac}}, \scalebox{0.8}{\cite{zhang2023act}},  \scalebox{0.8}{\cite{xie2024decomposition}}, \scalebox{0.8}{\cite{ren2024purple}}, \scalebox{0.8}{\cite{wang2024dac}}, \scalebox{0.8}{\cite{xie2024mag}}, \scalebox{0.8}{\cite{shen2024improving}}, \scalebox{0.8}{\cite{xu2024tcsr}}, \scalebox{0.8}{\cite{talaei2024chess}}, \scalebox{0.8}{\cite{qu2024before}}, \\ \scalebox{0.8}{\cite{lee2024mcs}},  \scalebox{0.8}{\cite{gorti2024msc}}, \scalebox{0.8}{\cite{liu2024epi}}, \scalebox{0.8}{\cite{thorpe2024dubo}}, \scalebox{0.8}{\cite{xia2024r3thissqlme}}, \scalebox{0.8}{\cite{wang2024improving}}, \scalebox{0.8}{\cite{mao2024enhancing}}, \scalebox{0.8}{\cite{tan2024enhancing}}, \scalebox{0.8}{\cite{shen2024selectsql}}, \scalebox{0.8}{\cite{luo2024ptdsql}}, \scalebox{0.8}{\cite{caferolu2024esql}}, \scalebox{0.8}{\cite{pourreza2024chasesql}}, \scalebox{0.8}{\cite{mai2024learning}}} \\ 
   \midrule
  \multirow{2}{*}{\scalebox{0.8}{RE}} & \scalebox{0.8}{Task Decomposition}  &  \scalebox{0.8}{36} & \makecell[l]{\scalebox{0.8}{\cite{pourreza2024din}}, \scalebox{0.8}{\cite{dong2023c3}}, \scalebox{0.8}{\cite{guo2023prompting}}, \scalebox{0.8}{\cite{sun2023sql}}, \scalebox{0.8}{\cite{sui2023reboost}}, \scalebox{0.8}{\cite{liu2023divide}}, \scalebox{0.8}{\cite{arora2023adapt}}, \scalebox{0.8}{\cite{guo2023retrieval}}, \scalebox{0.8}{\cite{gu2023interleaving}}, \scalebox{0.8}{\cite{ni2023lever}}, \scalebox{0.8}{\cite{wang2024mac}}, \scalebox{0.8}{\cite{ren2024purple}}, \scalebox{0.8}{\cite{fan2024metasql}}, \scalebox{0.8}{\cite{xie2024mag}}, \\ \scalebox{0.8}{\cite{xu2024tcsr}}, \scalebox{0.8}{\cite{cen2024sqlfix}}, \scalebox{0.8}{\cite{askari2024magic}}, \scalebox{0.8}{\cite{talaei2024chess}}, \scalebox{0.8}{\cite{qu2024before}}, \scalebox{0.8}{\cite{lee2024mcs}}, \scalebox{0.8}{\cite{volvovsky2024dfin}}, \scalebox{0.8}{\cite{mao2024enhancing}}, \scalebox{0.8}{\cite{wang2024tool}}, \scalebox{0.8}{\cite{luo2024ptdsql}}, \scalebox{0.8}{\cite{caferolu2024esql}}, \scalebox{0.8}{\cite{pourreza2024chasesql}}, \scalebox{0.8}{\cite{xie2024decomposition}}, \scalebox{0.8}{\cite{wang2024dac}}, \\ \scalebox{0.8}{\cite{li2024pet}}, \scalebox{0.8}{\cite{wang2023dbcopilot}}, \scalebox{0.8}{\cite{tan2024enhancing}}, \scalebox{0.8}{\cite{pourreza2024dts}}, \scalebox{0.8}{\cite{li2024sea}}, \scalebox{0.8}{\cite{zhang2024sqlfuse}}, \scalebox{0.8}{\cite{zhang2024finsql}}, \scalebox{0.8}{\cite{gorti2024msc}}} \\  
   & \scalebox{0.8}{Question Decomposition} & \scalebox{0.8}{10} & \scalebox{0.8}{\cite{arora2023adapt}}, \scalebox{0.8}{\cite{zhang2023act}}, \scalebox{0.8}{\cite{wang2024mac}}, \scalebox{0.8}{\cite{xie2024mag}}, \scalebox{0.8}{\cite{zhang2024structure}}, \scalebox{0.8}{\cite{fan2024metasql}}, \scalebox{0.8}{\cite{pourreza2024chasesql}}, \scalebox{0.8}{\cite{luo2024ptdsql}}, \scalebox{0.8}{\cite{shen2024selectsql}}, \scalebox{0.8}{\cite{tai2023exploring}}\\  
  \bottomrule
  \end{tabular}
  \end{center}
\end{table*}

\subsection{RQ2.3: What methodologies have been explored in the FT paradigm for LLM-based text-to-SQL?}
Compared to ICL-based approaches, FT-based methods for text-to-SQL tasks remain relatively underexplored, underscoring the need for further investigation into their potential. Recent studies have progressively proposed advanced FT-based techniques, which can be broadly grouped into two areas: \textit{Novel Model Architectures} and \textit{Model Training Strategies}. A summary of FT-based methods is listed in Table \ref{FT}.

\textbf{Novel Model Architectures.} In the FT paradigm, the development of innovative model architectures has become increasingly valuable in addressing specific challenges in SQL generation tasks \cite{kou2024cllms, pourreza2024sql, huang2024ccoe, ugare2024itergen, lin2024momq, zhong2024learning}. For instance, SQL generation using LLMs is significantly slower than traditional language modeling approaches \cite{li2023resdsql, yang2024harnessing}. Accordingly, CLLMs \cite{kou2024cllms} introduced a novel family of LLMs utilizing the Jacobi decoding method \cite{santilli2023accelerating} for faster SQL generation. CCoE \cite{huang2024ccoe} improved SQL generation performance by integrating multiple specialized domain experts into a single LLM, each trained individually to reduce resource requirements. ITERGEN \cite{ugare2024itergen} employed a novel grammar-guided technique with a symbol-to-position mapping to promote efficient and structured SQL generation. For real-world deployment scenarios requiring lightweight LLMs, KID \cite{zhong2024learning} introduced an innovative knowledge distillation approach, which transfers knowledge from a larger teacher model to a more compact student model. In response to the practical need for multi-dialect SQL generation, an MoE-based framework was introduced in MoMQ \cite{lin2024momq} for seamless cross-dialect querying in database management services. Similarly, SQL-GEN \cite{pourreza2024sql} enhanced model performance by employing a Mixture of Experts (MoE) initialization technique, thereby boosting model effectiveness across diverse SQL dialects.

\textbf{Model Training Strategies.} Two prominent fine-tuning approaches were commonly employed in LLM-based SQL generation, namely \textbf{Fully Fine-Tuning (FFT)} and \textbf{Parameter-Efficient Fine-Tuning (PEFT)}. While both methods leverage open-source LLMs for task-specific tuning, they differ significantly in their adaptation techniques. Specifically, FFT adjusts all parameters of open-source LLMs during fine-tuning \cite{wang2024mac, li2024codes, pourreza2024dts, pourreza2024sql, huang2024ccoe}, aiming to optimize the entire model based on relevant training data. This comprehensive adjustment often results in a more specialized model, capable of learning nuanced representations aligned with the specific SQL generation needs. It is particularly useful when the model requires significant modification to perform effectively on a new domain. In contrast, PEFT improves efficiency by adjusting only a subset of model parameters, making it less computationally intensive and reducing storage requirements compared to FFT \cite{fu2023effectiveness}. This approach is particularly ideal for resource-constrained environments, achieving competitive performance with minimal parameter updates. Among the commonly used PEFT techniques, \textit{Low-Rank Adaptation (LoRA)} \cite{hu2021lora} introduces low-rank updates to the model’s weights, while \textit{Quantized Low-Rank Adaptation (QLoRA)} \cite{dettmers2024qlora} further extends LoRA for models with quantized weights. Notably, LoRA and QLoRA have demonstrated significant training efficiency in text-to-SQL tasks \cite{zhou2024db2, zhang2024finsql, li2024sea, gorti2024msc, zhong2024learning}.

\begin{table*}[htbp] 
\caption{A summary of LLM-based methods employed in the FT paradigm for text-to-SQL tasks. These methods are categorized based on their specific implementation designs. Note that NMA-Novel Model Architecture, MTS-Model Training Strategy.}
\label{FT}
  \begin{center}
  \begin{tabular}{c|c|c|l}
  \toprule
  \scalebox{0.8}{\textbf{Category}} & \scalebox{0.8}{\textbf{Type}} & \scalebox{0.8}{\textbf{\# Studies}} & \scalebox{0.8}{\textbf{References}} \\
  \midrule
  \scalebox{0.8}{NMA} & - & \scalebox{0.8}{6} & \scalebox{0.8}{\cite{kou2024cllms}}, \scalebox{0.8}{\cite{pourreza2024sql}}, \scalebox{0.8}{\cite{huang2024ccoe}}, \scalebox{0.8}{\cite{ugare2024itergen}}, \scalebox{0.8}{\cite{lin2024momq}}, \scalebox{0.8}{\cite{zhong2024learning}} \\ 
  \midrule
  \multirow{2}{*}{\scalebox{0.8}{MTS}} & \scalebox{0.8}{FFT} & \scalebox{0.8}{11} & \scalebox{0.8}{\cite{li2024codes}}, \scalebox{0.8}{\cite{pourreza2024dts}}, \scalebox{0.8}{\cite{pourreza2024sql}}, \scalebox{0.8}{\cite{yang2024synthesizing}}, \scalebox{0.8}{\cite{huang2024ccoe}}, \scalebox{0.8}{\cite{xu2023symbol}}, \scalebox{0.8}{\cite{kou2024cllms}}, \scalebox{0.8}{\cite{zhuang2024structlm}}, \scalebox{0.8}{\cite{kobayashi2024yoro}}, \scalebox{0.8}{\cite{wu2024datagptsql}}, \scalebox{0.8}{\cite{ugare2024itergen}} \\ 
   & \scalebox{0.8}{PEFT} & \scalebox{0.8}{8} & \scalebox{0.8}{\cite{li2024sea}}, \scalebox{0.8}{\cite{zhang2024sqlfuse}}, \scalebox{0.8}{\cite{zhou2024db2}}, \scalebox{0.8}{\cite{chen2024open}}, \scalebox{0.8}{\cite{zhang2024finsql}}, \scalebox{0.8}{\cite{gorti2024msc}}, \scalebox{0.8}{\cite{lin2024momq}, \cite{zhong2024learning}} \\
  \bottomrule
  \end{tabular}
  \end{center}
\end{table*}

\begin{table*}[htbp] 
\caption{A summary of LLM-based methods employed in the post-processing paradigm for text-to-SQL tasks. These methods are categorized based on their specific implementation designs. Note that DSC-Direct SQL Correction, FGSC-Feedback-Guided SQL Correction, OC-Output Consistency, SC-Self Consistency, CC-Cross Consistency.}
\label{post}
  \begin{center}
  \begin{tabular}{c|c|c|l}
  \toprule
  \scalebox{0.8}{\textbf{Category}} & \scalebox{0.8}{\textbf{Type}} & \scalebox{0.8}{\textbf{\# Studies}} & \scalebox{0.8}{\textbf{References}} \\
  \midrule
  \scalebox{0.8}{DSC} & - & \scalebox{0.8}{10} & \scalebox{0.8}{\cite{pourreza2024din}}, \scalebox{0.8}{\cite{xie2024decomposition}}, \scalebox{0.8}{\cite{liu2023divide}}, \scalebox{0.8}{\cite{ren2024purple}}, \scalebox{0.8}{\cite{volvovsky2024dfin}}, \scalebox{0.8}{\cite{xu2024tcsr}}, \scalebox{0.8}{\cite{wang2024large}}, \scalebox{0.8}{\cite{caferolu2024esql}}, \scalebox{0.8}{\cite{zheng2024actor}}, \scalebox{0.8}{\cite{zhang2024finsql}} \\  
  \scalebox{0.8}{FGSC} & - & \scalebox{0.8}{27} & \makecell[l]{\scalebox{0.8}{\cite{guo2023prompting}}, \scalebox{0.8}{\cite{sun2023sql}}, \scalebox{0.8}{\cite{sui2023reboost}}, \scalebox{0.8}{\cite{ni2023lever}}, \scalebox{0.8}{\cite{guo2023retrieval}}, \scalebox{0.8}{\cite{gu2023interleaving}}, \scalebox{0.8}{\cite{wang2024mac}}, \scalebox{0.8}{\cite{wang2024dac}}, \scalebox{0.8}{\cite{xie2024mag}}, \scalebox{0.8}{\cite{gu2024middleware}}, \scalebox{0.8}{\cite{wu2024need}}, \scalebox{0.8}{\cite{cen2024sqlfix}}, \scalebox{0.8}{\cite{askari2024magic}}, \scalebox{0.8}{\cite{talaei2024chess}}, \scalebox{0.8}{\cite{yan2024intelliexplain}}, \scalebox{0.8}{\cite{liu2024epi}}, \scalebox{0.8}{\cite{thorpe2024dubo}}, \\\scalebox{0.8}{\cite{chen2024teaching}}, \scalebox{0.8}{\cite{xia2024r3thissqlme}}, \scalebox{0.8}{\cite{mao2024enhancing}}, \scalebox{0.8}{\cite{wang2024tool}}, \scalebox{0.8}{\cite{shen2024selectsql}}, \scalebox{0.8}{\cite{pourreza2024chasesql}}, \scalebox{0.8}{\cite{zheng2024actor}}, \scalebox{0.8}{\cite{li2024sea}}, \scalebox{0.8}{\cite{wu2024datagptsql}}, \scalebox{0.8}{\cite{zhang2024sqlfuse}}     } \\
  \midrule
  \multirow{2}{*}{\scalebox{0.8}{OC}} & \scalebox{0.8}{SC} & \scalebox{0.8}{13} & \scalebox{0.8}{\cite{dong2023c3}}, \scalebox{0.8}{\cite{gao2023text}}, \scalebox{0.8}{\cite{sun2023sqlprompt}}, \scalebox{0.8}{\cite{ni2023lever}}, \scalebox{0.8}{\cite{ren2024purple}}, \scalebox{0.8}{\cite{fan2024metasql}}, \scalebox{0.8}{\cite{lee2024mcs}}, \scalebox{0.8}{\cite{li2024using}}, \scalebox{0.8}{\cite{tan2024enhancing}}, \scalebox{0.8}{\cite{shen2024selectsql}}, \scalebox{0.8}{\cite{pourreza2024chasesql}}, \scalebox{0.8}{\cite{zhang2024finsql}}, \scalebox{0.8}{\cite{gorti2024msc}} \\  
   & \scalebox{0.8}{CC} & \scalebox{0.8}{1} & \scalebox{0.8}{\cite{li2024pet}} \\
  \bottomrule
  \end{tabular}
  \end{center}
\end{table*}

\subsection{RQ2.4: What methodologies have been explored in the post-processing paradigm for LLM-based text-to-SQL?}
After producing SQL queries, text-to-SQL systems often require further refinement to align outputs more closely with user expectations. As illustrated in Table \ref{post}, current post-processing approaches are categorized into \textit{Direct SQL Correction}, \textit{Feedback-Guided SQL Correction}, and \textit{Output Consistency}.

\textbf{Direct SQL Correction.} SQL queries produced by LLMs may contain syntax errors, necessitating direct SQL correction mechanisms to improve their quality \cite{pourreza2024din, ren2024purple, xu2024tcsr, liu2023divide, wang2024large, caferolu2024esql}. Several methods have been developed to address these issues. For instance, DIN-SQL \cite{pourreza2024din}, TCSR-SQL \cite{xu2024tcsr}, and E-SQL \cite{caferolu2024esql} integrated self-correction guidelines that enable LLMs to detect and rectify syntactic errors. PURPLE \cite{ren2024purple} categorized six primary types of SQL errors typically made by LLMs and offered targeted error corrections. GR-DnP \cite{liu2023divide} generated an initial SQL draft, followed by the refinement based on the generated SQL. MLPrompt \cite{wang2024large} automatically translated some predefined rules that an LLM struggled to follow into other languages, thereby drawing greater attention to those rules. Although advances in LLMs are expected to reduce syntax errors, prompting models to autonomously identify and correct these issues remains a promising strategy to improve SQL quality.

\textbf{Feedback-Guided SQL Correction.} In text-to-SQL tasks, incorporating feedback into post-processing can significantly improve SQL accuracy by guiding necessary refinements. Several models have leveraged feedback to enhance SQL generation \cite{wang2024mac, talaei2024chess, chen2024teaching, hong2024knowledge, guo2023prompting}. For instance, MAC-SQL \cite{wang2024mac} introduced a Refiner Agent that automatically detected and corrected SQL errors by executing SQL, gathering feedback, and iteratively adjusting problematic SQL. CHESS \cite{talaei2024chess} initiated SQL generation with a draft SQL, which was iteratively refined based on execution feedback to improve accuracy. SELF-DEBUGGING \cite{chen2024teaching} trained LLMs to autonomously debug SQL queries. Using few-shot learning, they analyzed execution results and provided NL explanations for each correction without human intervention. Knowledge-to-SQL \cite{hong2024knowledge} introduced Direct Preference Optimization (DPO) \cite{rafailov2024direct} to optimize SQL by incorporating execution feedback, whereas DESEM \cite{guo2023prompting} designed a fallback revision mechanism to refine SQL queries based on execution errors. It also included termination criteria to prevent excessive corrections. 

\textbf{Output Consistency.} To strengthen the consistency of model outputs in text-to-SQL tasks, two main approaches have been introduced in the existing literature: \textit{self-consistency} and \textit{cross-consistency}. Specifically, self-consistency involves two primary strategies: (1) Multiple reasoning paths are generated within a single LLM, and the most consistent SQL is selected \cite{dong2023c3, gao2023text, fan2024metasql, ren2024purple, talaei2024chess}; (2) LLMs are employed as verifiers to identify the most appropriate SQL from multiple candidates \cite{ni2023lever, li2024using, lee2024mcs}. For instance, C3 \cite{dong2023c3}, DAIL-SQL \cite{gao2023text}, MetaSQL \cite{fan2024metasql}, PURPLE \cite{ren2024purple}, and CHESS \cite{talaei2024chess} reduced output noise by selecting the most consistent SQL query among several generated options. Additionally, LEVER \cite{ni2023lever} and \cite{li2024using} incorporated re-ranking techniques to select the optimal SQL from multiple candidates. MCS-SQL \cite{lee2024mcs} first generated various candidate SQL using diverse prompts, followed by a filtering process and a multiple-choice selection process to choose the final SQL. In contrast, cross-consistency involves guiding various LLMs to produce SQL queries collaboratively and conduct voting mechanisms to identify the best outcome. Notably, PET-SQL \cite{li2024pet} applied cross-consistency in generating diverse SQL queries from various LLMs, thereby ensuring higher accuracy and consistency.

\begin{tcolorbox}[colback=black!5!white, colframe=black, 
                  title=\textcolor{black}{\textbf{Answer to RQ2}}, 
                  coltitle=white, colbacktitle=pink, 
                  boxrule=1pt, width=\textwidth]
(1) In the pre-processing paradigm, schema linking is the most widely used technique for connecting NL queries directly to database tables and columns. Moreover, cell value acquisition focuses on retrieving relevant cell values from the database schema, while question rewriting helps mitigate ambiguities in user queries. Besides, data augmentation can be used to enlarge training data and improve overall data quality. 

(2) The ICL paradigm can be divided into three main approaches: supplementary knowledge incorporation, example selection, and reasoning enhancement. In particular, SQL knowledge and question knowledge are often incorporated as supplementary knowledge to facilitate SQL generation. Moreover, few-shot prompting is more widely used than zero-shot prompting, and it generally yields more reliable results. For reasoning enhancement, task composition and question decomposition are often utilized in complicated and multi-step SQL generation.

(3) In the FT paradigm, novel model architectures remain limited, with only six surveyed papers in current research, highlighting an area for further exploration. Moreover, most FT-based methods utilize FFT rather than PEFT, and FFT typically produces better performance outcomes.

(4) In the post-processing paradigm, feedback-guided SQL correction is the most commonly used strategy, which incorporates feedback to support automatic SQL correction. Moreover, output consistency also plays a critical role, aiming to increase the accuracy and reliability of the generated SQL. Notably, self-consistency is more prevalent than cross-consistency. In contrast, direct SQL correction is less widely adopted, since advancements in LLMs are expected to reduce SQL syntax errors over time.

\end{tcolorbox}

\section{RQ3: What are the characteristics of the datasets and evaluation metrics?} \label{RQ3}
In this research question, we summarize the existing datasets and evaluation metrics in text-to-SQL tasks.

\begin{table*}[ht]
\centering
\caption{A summary of text-to-SQL datasets, including dataset size, dataset turn, dataset type, and dataset release year. Note that these datasets are all open to the public.}
\label{dataset}
\begin{tabular}{c|c|c|c}
\hline
{\scalebox{0.8}{\textbf{Dataset Name}}}  & {\scalebox{0.8}{\textbf{Dataset Turn}}} & {\scalebox{0.8}{\textbf{Dataset Type}}} & {\scalebox{0.8}{\textbf{Release Year}}} \\
\hline
\scalebox{0.8}{MIMICSQL \cite{wang2020text}} & \scalebox{0.8}{Single-Turn} & \scalebox{0.8}{Single-Domain} & \scalebox{0.8}{2020} \\
\scalebox{0.8}{EHRSQL \cite{lee2022ehrsql}} & \scalebox{0.8}{Single-Turn} & \scalebox{0.8}{Single-Domain} & \scalebox{0.8}{2022} \\ 
\scalebox{0.8}{ScienceBenchmark \cite{zhang2023sciencebenchmark}} & \scalebox{0.8}{Single-Turn} & \scalebox{0.8}{Single-Domain} & \scalebox{0.8}{2023} \\
\scalebox{0.8}{BookSQL \cite{kumar2024booksql}} & \scalebox{0.8}{Single-Turn} & \scalebox{0.8}{Single-Domain} & \scalebox{0.8}{2024} \\
\scalebox{0.8}{WikiSQL \cite{zhong2017seq2sql}} & \scalebox{0.8}{Single-Turn} & \scalebox{0.8}{Cross-Domain} & \scalebox{0.8}{2017} \\
\scalebox{0.8}{Spider \cite{yu2018spider}} & \scalebox{0.8}{Single-Turn} & \scalebox{0.8}{Cross-Domain} & \scalebox{0.8}{2018} \\
\scalebox{0.8}{KaggleDBQA \cite{lee2021kaggledbqa}} & \scalebox{0.8}{Single-Turn} & \scalebox{0.8}{Cross-Domain} & \scalebox{0.8}{2021} \\
\scalebox{0.8}{BIRD \cite{li2024can}} & \scalebox{0.8}{Single-Turn} & \scalebox{0.8}{Cross-Domain} & \scalebox{0.8}{2024} \\
\scalebox{0.8}{BEAVER \cite{chen2024beaver}} & \scalebox{0.8}{Single-Turn} & \scalebox{0.8}{Cross-Domain} & \scalebox{0.8}{2024} \\
\scalebox{0.8}{CSpider \cite{min2019pilot}} & \scalebox{0.8}{Single-Turn} & \scalebox{0.8}{Cross-Domain (Cross-Lingual)} & \scalebox{0.8}{2019} \\
\scalebox{0.8}{DuSQL \cite{wang2020dusql}} & \scalebox{0.8}{Multi-Turn} & \scalebox{0.8}{Cross-Domain (Cross-Lingual)} & \scalebox{0.8}{2020} \\
\scalebox{0.8}{CHASE \cite{guo2021chase}} & \scalebox{0.8}{Multi-Turn} & \scalebox{0.8}{Cross-Domain (Cross-Lingual)} & \scalebox{0.8}{2021} \\
\scalebox{0.8}{Multi-Spider \cite{dou2023multispider}} & \scalebox{0.8}{Single-Turn} & \scalebox{0.8}{Cross-Domain (Cross-Lingual)} & \scalebox{0.8}{2023} \\
\scalebox{0.8}{StatBot.Swiss \cite{nooralahzadeh2024statbot}} & \scalebox{0.8}{Single-Turn} & \scalebox{0.8}{Cross-Domain (Cross-Lingual)} & \scalebox{0.8}{2024} \\
\scalebox{0.8}{SParC \cite{yu2019sparc}} & \scalebox{0.8}{Multi-Turn} & \scalebox{0.8}{Cross-Domain} & \scalebox{0.8}{2019} \\
\scalebox{0.8}{CoSQL \cite{yu2019cosql}} & \scalebox{0.8}{Multi-Turn} & \scalebox{0.8}{Cross-Domain} & \scalebox{0.8}{2019} \\
\scalebox{0.8}{EHR-SeqSQL \cite{ryu2024ehr}} & \scalebox{0.8}{Multi-Turn} & \scalebox{0.8}{Cross-Domain} & \scalebox{0.8}{2024} \\
\scalebox{0.8}{Spider-Realistic \cite{deng2020structure}} & \scalebox{0.8}{Single-Turn} & \scalebox{0.8}{Cross-Domain (Robustness-Centered)} & \scalebox{0.8}{2020} \\
\scalebox{0.8}{Spider-Syn \cite{gan2021towards}} & \scalebox{0.8}{Single-Turn} & \scalebox{0.8}{Cross-Domain (Robustness-Centered)} & \scalebox{0.8}{2021} \\
\scalebox{0.8}{Spider-DK \cite{gan2021exploring}} & \scalebox{0.8}{Single-Turn} & \scalebox{0.8}{Cross-Domain (Robustness-Centered)} & \scalebox{0.8}{2021} \\
\scalebox{0.8}{Spider-SSP \cite{shaw2021compositional}} & \scalebox{0.8}{Single-Turn} & \scalebox{0.8}{Cross-Domain (Robustness-Centered)} & \scalebox{0.8}{2021} \\
\scalebox{0.8}{Spider-SS\&CG \cite{gan2022measuring}} & \scalebox{0.8}{Single-Turn} & \scalebox{0.8}{Cross-Domain (Robustness-Centered)} & \scalebox{0.8}{2022} \\
\scalebox{0.8}{Spider-GEN \cite{patil2023exploring}} & \scalebox{0.8}{Single-Turn} & \scalebox{0.8}{Cross-Domain (Robustness-Centered)} & \scalebox{0.8}{2022} \\
\scalebox{0.8}{Dr.Spider \cite{chang2023dr}} & \scalebox{0.8}{Single-Turn} & \scalebox{0.8}{Cross-Domain (Robustness-Centered)} & \scalebox{0.8}{2023} \\
\scalebox{0.8}{AmbiQT \cite{bhaskar2023benchmarking}} & \scalebox{0.8}{Single-Turn} & \scalebox{0.8}{Cross-Domain (Robustness-Centered)} & \scalebox{0.8}{2023} \\
\scalebox{0.8}{AMBROSIA \cite{saparina2024ambrosia}} & \scalebox{0.8}{Single-Turn} & \scalebox{0.8}{Cross-Domain (Robustness-Centered)} & \scalebox{0.8}{2024} \\

\bottomrule
\end{tabular}
\end{table*}

\subsection{RQ3.1: What are the existing datasets in text-to-SQL tasks?}
Amidst the rapid advancements in the text-to-SQL domain, numerous datasets have been developed to evaluate model performance in SQL generation. In this research question, we roughly categorize several widely used datasets into the following groups. A summary of these datasets is provided in Table \ref{dataset}.

\textbf{Single-Domain Text-to-SQL Datasets.} Large-scale single-domain datasets, such as MIMICSQL \cite{wang2020text}, EHRSQL \cite{lee2022ehrsql}, ScienceBenchmark \cite{zhang2023sciencebenchmark}, and BookSQL \cite{kumar2024booksql}, are primarily designed for specific domain applications. These datasets contain intricate databases and SQL queries tailored to specific applications, reflecting an increasing emphasis on assessing text-to-SQL systems within specialized domains.

\textbf{Cross-Domain Text-to-SQL Datasets.} Cross-domain datasets are designed to gauge the generalization abilities of text-to-SQL systems across various domains. These datasets require models to effectively adapt to unfamiliar SQL queries and databases. They can be further categorized as follows:

(1) \textbf{Ordinary Text-to-SQL Datasets.} Ordinary cross-domain text-to-SQL datasets typically do not consider other important features, such as cross-lingual support, multi-turn interactions, or robustness to the changing external environments. For instance, WikiSQL \cite{zhong2017seq2sql} comprises 80,654 manually curated pairs of NL questions and SQL queries, each associated with a unique table. Subsequently, Spider \cite{yu2018spider} introduces 11,840 NL questions and 6,448 unique SQL queries across 138 distinct domains, featuring databases with complex table relationships. KaggleDBQA \cite{lee2021kaggledbqa} enhances text-to-SQL systems by providing comprehensive database schema information, including column and table descriptions, as well as explanations of categorical values. More recently, BIRD \cite{li2024can} has gained widespread attention for incorporating additional complexities, such as complex SQL functions and operations, further challenging the model generalization capabilities of SQL generation.

(2) \textbf{Cross-Lingual Text-to-SQL Datasets.} SQL keywords, along with table and column names, are predominantly in English, which poses challenges for scenarios targeting non-English languages. This linguistic mismatch complicates question understanding and SQL generation. For instance, CSpider \cite{min2019pilot} translates the original Spider dataset into Chinese, highlighting difficulties such as word segmentation and cross-lingual mapping due to discrepancies in grammatical structures. DuSQL \cite{wang2020dusql} and CHASE \cite{guo2021chase} provide context-dependent text-to-SQL datasets featuring questions and database content in Chinese and English, which are used to address practical challenges in multilingual scenarios. MultiSpider \cite{dou2023multispider} introduces lexical and structural complexities by incorporating seven different languages, whereas StatBot.Swiss \cite{nooralahzadeh2024statbot} offers a more realistic and intricate dataset featuring both English and German languages.

(3) \textbf{Context-Dependent Text-to-SQL Datasets.} There has been a growing emphasis on developing text-to-SQL systems capable of handling multi-round conversations. As a result, several context-dependent datasets have been introduced. For instance, SParC \cite{yu2019sparc} provides over 12,000 sequentially linked (NL, SQL) pairs across various domains, requiring text-to-SQL systems to understand context throughout dialogues. CoSQL \cite{yu2019cosql} offers more than 30,000 conversational turns, presenting challenges like addressing unanswerable user queries and comprehending context across multiple interactions. DuSQL \cite{wang2020dusql} and CHASE \cite{guo2021chase} reduce the number of simple SQL queries from CoSQL, further testing the capabilities of dialogue-centered text-to-SQL systems. As the first multi-turn medical text-to-SQL dataset, EHR-SeqSQL \cite{ryu2024ehr} features sequential questions and integrates specially designed tokens into SQL queries to facilitate SQL generation. These datasets highlight the need for more complex and context-dependent interactions, thereby enhancing the model's capacity to manage dynamic and conversational text-to-SQL tasks.

(4) \textbf{Robustness-Centered Text-to-SQL Datasets.} In practical applications, text-to-SQL systems need to cope with diverse user queries and function effectively across multiple databases, showcasing the significance of robust system design. Consequently, several datasets have been developed to assess and enhance the robustness of text-to-SQL systems. For instance, Spider-DK \cite{gan2021exploring} integrates five categories of domain knowledge into NL questions to evaluate the system's ability to utilize this information. Spider-Realistic \cite{deng2020structure} presents 508 complex cases with rephrased NL questions, further testing the system's ability to handle subtle linguistic variations. Spider-Syn \cite{gan2021towards} emulates situations where users might be unfamiliar with the database by substituting database-related terms in NL questions with synonyms.  Spider-SSP \cite{shaw2021compositional} centers on schema-oriented semantic parsing techniques to evaluate schema generalization by modifying column and table names within the database. Spider-SS\&CG \cite{gan2022measuring} focuses on assessing task performance in real databases by simplifying database schema and increasing database structure complexity. Dr.Spider \cite{chang2023dr} introduces 17 different types of modifications to the databases, NL questions, and SQL queries from the original Spider dataset, offering a more comprehensive benchmark for evaluating the robustness of text-to-SQL systems in real-world scenarios. Additionally, from the perspective of question ambiguity, AmbiQT \cite{bhaskar2023benchmarking} is designated to gauge the capability of text-to-SQL systems to handle ambiguity, featuring over 3000 cases that highlight four types of ambiguous scenarios. AMBROSIA \cite{saparina2024ambrosia} includes questions that primarily exhibit three types of ambiguities along with their interpretations and corresponding SQL queries.

\subsection{RQ3.2: What are the commonly used evaluation metrics in text-to-SQL tasks?}
Evaluation metrics are crucial for gauging the efficacy of text-to-SQL systems from a quantitative perspective, which can be divided into content matching-based metrics and execution result-based metrics. In the following content, $N$ indicates the dataset size, $Q_i$ denotes the $i$-th NL question, $R_i$ represents the $i$-th execution result from the ground-truth SQL $S_i$, and $\hat{R}_i$ refers to the $i$-th execution result from the generated SQL $\hat{S}_i$.

\textbf{Content Matching-based Metrics.} Content matching-based metrics concentrate on evaluating the similarity between the produced and the ground-truth SQL through structural and syntactical comparisons.

(1) \textbf{String-Match Accuracy (SM)} \cite{zhong2017seq2sql} is introduced to evaluate the model's effectiveness of SQL generation by determining whether the produced SQL exactly matches its corresponding ground truth, without considering execution results or functional equivalence. This strict metric only rewards models when there is a completely identical match, making it particularly effective for assessing syntactical correctness. However, it may overlook cases where semantically equivalent queries are expressed in different ways. It can be computed as:
\begin{equation}
    \mathrm{SM}=\frac{\sum_{i=1}^N\mathbf{1}(S_i=\hat{S_i})}{N}
\end{equation}
where $\mathbf{1}(\cdot)$ denotes an indicator function that equals to 1 if the condition inside is satisfied, and 0 otherwise.

(2) \textbf{Component-Match Accuracy (CM)} \cite{yu2018spider} assesses the efficacy of text-to-SQL systems via examining the alignment of specific SQL components in the predicted SQL with their corresponding ground-truth counterparts. In contrast to SM, CM provides a more comprehensive assessment by awarding credit for partially correct SQL and calculating the accuracy of individual SQL components. This approach provides a more nuanced understanding of the model's performance in generating SQL queries. The computation for a specific SQL component $C_k$ can be expressed as:
\begin{equation}
    \mathrm{CM}=\frac{\sum_{i=1}^N\mathbf{1}(S_i^{C_k}=\hat{S}_i^{C_k})}{N}
\end{equation}
where $S_i^{C_k}$ and $\hat{S}_i^{C_k}$ are one of the components of SQL queries $S_i$ and $\hat{S}_i$, respectively. 

(3) \textbf{Exact-Match Accuracy (EM)} \cite{yu2018spider} builds upon CM by evaluating whether every component $C_k$ in the predicted SQL precisely matches the counterpart from the ground truths. It provides a holistic measure of SQL correctness by requiring that every SQL component aligns exactly. As a result, EM is a stricter metric compared to CM, ensuring syntactical and structural alignment in the evaluation of SQL queries. It can be formulated as:
\begin{equation}
    \mathrm{EM}=\frac{\sum_{i=1}^N\mathbf{1}(\bigcap_{C_k\in\mathbb{C}}S_i^{C_k}=\hat{S}_i^{C_k})}{N}
\end{equation}
where $\bigcap_{C_k\in\mathbb{C}}S_i^{C_k}=\hat{S}_i^{C_k}$ implies that every $C_k$ in the generated SQL is completely identical to the corresponding component in the ground truth.

(4) \textbf{SQL Query Analysis Metric (SQAM)} \cite{song2024enhancing} is introduced to assess the efficacy of SQL generation by decomposing it into several keywords like SELECT, WHERE, ORDER BY, etc. These components are further segmented into sub-components based on different entities within each category, such as selected table and column names. Finally, SQAM is calculated based on the degree of alignment among these sub-components, with a value ranging from 0 to 1.

(5) \textbf{Tree Similarity of Editing Distance (TSED)} \cite{song2024enhancing} is specially designed to employ abstract syntax tree (AST) as features, measuring the discrepancy between two ASTs derived from the generated SQL and its corresponding ground truth. Subsequently, TSED is computed using the discrepancy $D$ and the node count of the larger AST $N$ from the predicted SQL or the ground truth. Mathematically, it can be given as:
\begin{equation}
    \mathrm{TSED}=\frac{D}{N}
\end{equation} 

(6) \textbf{ESM+} \cite{ascoli2024esm+} is a recently introduced metric that extends EM by incorporating additional rules for handling specific SQL keywords, such as JOIN, DISTINCT, LIMIT, etc. This enhancement aims to provide a more accurate evaluation by minimizing false positives and false negatives frequently observed in the original EM.

\textbf{Execution Result-based Metrics.} Execution result-based metrics gauge the validity of the produced SQL by comparing the results from SQL execution with the anticipated outcomes. These metrics prioritize whether the SQL produces the correct results, reflecting its functional accuracy rather than merely its syntactical or structural correctness. Accordingly, they offer a more practical assessment of the system's performance in real-world scenarios.

(1) \textbf{Execution Accuracy (EX)} \cite{yu2018spider} assesses the performance of text-to-SQL systems by comparing the execution results of the generated SQL with the expected outcomes. This metric is valuable for verifying the functional correctness of SQL queries, which can be given by:
\begin{equation}
    \mathrm{EX}=\frac{\sum_{i=1}^N\mathbf{1}(R_i=\hat{R}_i)}{N}
\end{equation}

(2) \textbf{Test-suite Accuracy (TS)} \cite{zhong2020semantic} is proposed to gauge the semantic correctness of text-to-SQL systems by constructing a compact test suite from a substantial collection of databases. This approach enables the differentiation between fully correct and nearly correct SQL queries. During the evaluation phase, TS measures the model’s ability to correctly execute SQL queries across these databases, thereby establishing a stringent upper bound for semantic accuracy. 

(3) \textbf{Valid Efficiency Score (VES)} \cite{li2024can} quantifies the execution efficiency of SQL queries by simultaneously considering the accuracy and execution efficiency of SQL outputs. It can be formally calculated as:
\begin{equation}
    \mathrm{VES}=\frac{\sum_{i=1}^N\mathbf{1}(R_i=\hat{R}_i)\cdot\mathbf{R}(S_i,\hat{S}_i)}{N}
\end{equation}
\begin{equation}
    \mathbf{R}(S_i,\hat{S}_i)=\sqrt{\frac{\mathbf{E}(S_i)}{\mathbf{E}(\hat{S}_i)}}
\end{equation}
where $\mathbf{R}(S_i,\hat{S}_i)$ evaluates the relative execution efficiency of the produced SQL against their ground truths. Meanwhile, $\mathbf{E}(\cdot)$ computes the efficiency of the SQL query, which may be assessed based on memory usage and execution time. 

(4) \textbf{Reward-based Valid Efficiency Score (R-VES)} \cite{li2024can} enhances VES to ensure greater stability and reliability. Rather than merely calculating the time ratio between the predicted and ground-truth SQL, R-VES incorporates reward points alongside this time ratio to provide a more nuanced assessment. This metric can be formulated as:
\begin{equation}
    \begin{aligned}&\text{R-VES}=\begin{cases}1.25&\text{if }\hat{S}_i \text{ is correct and } \tau\geq2\\1&\text{if }\hat{S}_i \text{ is correct and } 1\leq\tau<2\\0.75&\text{if }\hat{y} \text{ is correct and } 0.5\leq\tau<1\\0.5&\text{if }\hat{S}_i \text{ is correct and } 0.25\leq\tau<0.5\\0.25&\text{if }\hat{S}_i\text{ is correct and } \tau<0.25\\0&\text{if }\hat{S}_i\text{ is incorrect}\end{cases}\end{aligned}
\end{equation}
where $\tau=\frac{\text{Ground truth SQL run time}}{\text{Predicted SQL run time}}$ stands for the time ratio, which is determined by executing SQL for 100 times, averaging the results, and removing any outliers.

(5) \textbf{Soft-F1} \cite{li2024can} is tailored to evaluate text-to-SQL model performance by calculating the similarity between the tables from the predicted SQL and those from the corresponding ground truths. This metric provides a more comprehensive assessment by mitigating the effects of column order and missing values in the tables of the generated SQL. 

(6) \textbf{Query Variance Testing (QVT)} \cite{li2024dawn} assesses the robustness of text-to-SQL systems by measuring the ability to generate correct SQL across diverse formulations from the same question. Specifically, $m_i$ NL question variations correspond to a given question $Q_i$ and a SQL query $S_i$, and this metric can be computed as:
\begin{equation}
    \mathrm{QVT}=\frac{1}{N}\sum_{i=1}^{N}\left(\frac{\sum_{j=1}^{m_i}\mathbf{1}\left(\mathcal{F}(Q_{ij})=S_i\right)}{m_i}\right)
\end{equation}
where $\mathcal{F}(Q_{ij})$ indicates the generated SQL for the $j$-th NL variation of $S_i$.

\begin{tcolorbox}[colback=black!5!white, colframe=black, 
                  title=\textcolor{black}{\textbf{Answer to RQ3}}, 
                  coltitle=white, colbacktitle=pink, 
                  boxrule=1pt, width=\textwidth]
(1) Existing text-to-SQL datasets can be broadly categorized into single-domain and cross-domain types. In particular, cross-domain datasets can be further divided into original cross-domain, cross-lingual, context-dependent, and robustness-centered datasets. These benchmarks are often used to evaluate model performance in SQL generation.

(2) Existing text-to-SQL evaluation metrics can be classified into content matching-based metrics (including SM, CM, EM, SQAM, TSED, ESM+) and execution result-based metrics (including EX, TS, VES, R-VES, Soft-F1, QVT). These metrics offer a systematic framework for assessing the performance of models in SQL generation.
\end{tcolorbox}

\section{RQ4: What are the challenges and future directions for text-to-SQL?} \label{RQ4}
Despite significant advancements in LLM-based text-to-SQL research, several challenges remain unresolved and demand further exploration. In this research question, some key obstacles and potential future directions are discussed in detail.

\subsection{RQ4.1: What are the key unresolved challenges in the surveyed studies?} 

\textbf{Challenge 1: Structural Complexity and Linguistic Ambiguity.} Accurately interpreting user intent and identifying key question components are crucial for effective text-to-SQL conversion. However, users often struggle to clearly articulate their questions in real-world applications, resulting in structural complexity and linguistic ambiguity \cite{kim2023tree}. These issues can be further categorized as follows: (1) \textbf{Structural Complexity}: NL queries often exhibit intricate structures like nested clauses, co-references, and ellipses, thereby complicating their direct mappings to SQL queries. Additionally, real-world inputs often include spelling and grammatical mistakes, further impeding precise NL interpretation. (2) \textbf{Lexical Ambiguity}: This issue arises when a term possesses several interpretations, potentially leading to misinterpretations of user intent. Resolving such ambiguity requires models to capture schema-specific details and leverage contextual information to determine the intended meaning of each term within the NL query. (3) \textbf{Syntactic Ambiguity}: NL queries are inherently ambiguous, often allowing for multiple plausible interpretations. This issue is particularly prevalent when the context is insufficient or when domain-specific knowledge is required to disambiguate different syntactic structures \cite{kim2024m}. Although question rewriting techniques can help mitigate these issues to some extent, current text-to-SQL approaches have not sufficiently addressed them, highlighting a crucial area for future research.

\textbf{Challenge 2: Insufficient Database Schema Comprehension.} Accurately comprehending complex database schema is essential for representing elements such as table and column names, along with inter-table relationships. However, current LLM-based approaches often struggle to fully capture relevant schema information across diverse domains, which can be attributed to the following aspects: (1) \textbf{Variations in Domain-Specific Schemas}: Different domains often adopt unique database design conventions, terminologies, and abbreviations. Such variations complicate the SQL generation process, since the model needs to accurately interpret the schema structure and domain-specific jargon. (2) \textbf{Similar Column Names Across Tables}: Databases frequently contain tables with similar column names serving distinct purposes. This can lead to confusion in SQL generation, since text-to-SQL systems need to correctly differentiate these columns based on the specific query context and user intent. (3) \textbf{Intricate Relationships Across Multiple Tables}: Complex databases often contain multiple tables with complex interrelationships, and accurately navigating these relationships is fundamental for generating valid SQL queries. Misunderstanding these connections may result in errors such as incorrect joins or missing data in the generated SQL.

\textbf{Challenge 3: Poor Cross-Domain Generalization.} A significant challenge in SQL generation is enabling models to generalize effectively across unseen databases and domains. Each domain often utilizes distinct data structures, relationships, and terminologies, which can hinder the model’s performance when confronted with databases outside of its training scope. For instance, many existing LLM-based models perform well on benchmarks like the Spider dataset but tend to struggle on others, such as BIRD or some robustness-focused datasets. This limited cross-domain adaptability restricts the real-world applicability of text-to-SQL systems, emphasizing the need for models capable of dynamically adjusting to new database schemas and diverse domain-specific contexts.

\textbf{Challenge 4: Lower Model Efficiency.} Model efficiency is critical for advancing text-to-SQL systems, especially as benchmarks grow more complex with additional tables and columns. This increasing complexity introduces the following challenges: (1) \textbf{Input Capacity Limitation.} Input capacity limitation of LLMs is one of the major issues, especially when database schema size exceeds the model's maximum token limits. This is particularly apparent in open-source LLMs with shorter context lengths \cite{guo2024deepseek}, which can lead to the exclusion of essential information and negatively undermine accurate SQL generation \cite{wang2024beyond}. (2) \textbf{Slower Inference Speed.} LLM-based methods generally exhibit slower inference speeds compared to pre-trained language model (PLM)-based methods \cite{li2023resdsql, yang2024harnessing}. A primary reason for this inefficiency is that many LLM-based approaches take the entire database schema as input, resulting in redundant data processing and higher computational costs \cite{wang2024mac}. Addressing these efficiency challenges is critical for strengthening the scalability and practical implementation of text-to-SQL systems.

\textbf{Challenge 5: Inadequate and Poor-Quality Training Data.} High-quality annotated data is essential for developing text-to-SQL systems, but acquiring such data remains challenging. Specifically, manually constructing accurate and diverse (NL, SQL) pairs necessitates substantial domain expertise, rendering data curation labor-intensive and prone to errors and inconsistencies. In particular, existing publicly available datasets like Spider and BIRD are often limited in size, domain diversity, and annotation quality, which hinder effective model generalization \cite{wretblad2024understanding}. These limitations in data availability and quality underscore the need for advanced data augmentation strategies to improve model performance.

\textbf{Challenge 6: Potential Data Privacy and Low Model Reliability.} Ensuring data privacy is essential, especially when handling sensitive databases. Nonetheless, recent studies predominantly utilize ICL-based methods, often relying on closed-source models for implementation \cite{pourreza2024din}. This reliance on proprietary LLMs poses a significant privacy risk, since sharing database schemas, sample data, or domain-specific knowledge with third-party API providers may compromise data security \cite{kibriya2024privacy}. Although FT-based approaches help alleviate this issue by keeping data on-premises, they often exhibit sub-optimal performance \cite{pourreza2024dts, li2024codes}. Moreover, the lack of interpretability in LLMs poses challenges for the development of text-to-SQL systems, which are crucial for several reasons. (1) It enables users to understand why the model generates specific SQL queries, which is essential for maintaining user trust in the system. (2) It offers insights into the decision-making process, allowing developers to identify and rectify potential biases or reasoning errors. However, interpretability remains an underexplored area in LLM-based SQL generation, highlighting an urgent need for models that balance performance with transparency and user trust.

\textbf{Challenge 7: Additional Technical Challenges in SQL Generation.} In addition to the above challenges, developing robust text-to-SQL systems also involves overcoming several technical hurdles: (1) \textbf{Unpredictable SQL Generation.} The outputs of ICL-based models can vary widely depending on the specific prompts. This unpredictability can lead to inconsistencies in the SQL queries generated for identical questions, which stems from the model's probabilistic nature, the nuanced structure of NL inputs, and differences in prompt interpretation. Such instability in output generation complicates the adoption of LLMs in practical applications. (2) \textbf{Inherent Model Biases.} Prompt-based models frequently exhibit biases that can lead to incorrect SQL outputs \cite{turpin2024language}. For instance, GPT-series models tend to convert uppercase values from NL queries to lowercase in the SQL outputs and often  prefer 'LEFT JOIN' over 'JOIN' \cite{liu2023comprehensive}. These biases can significantly affect precise SQL generation when faced with complex database schemas. (3) \textbf{Model Hallucination.} Model hallucination poses a unique challenge in text-to-SQL tasks, since it can result in producing incorrect SQL queries from NL inputs. Particularly, model hallucination occurs when the SQL queries produced by LLMs fail to accurately reflect the user's intent or the database schemas. For instance, LLMs may select nonexistent tables or columns, use incorrect SQL functions, or misinterpret table relationships. Consequently, addressing the above issues is critical for improving the correctness of LLM-generated SQL queries.

\subsection{RQ4.2: What are potential future directions for advancing LLM-based text-to-SQL systems?} \label{future}

\textbf{Direction 1: Alleviating Question Ambiguity.} To address ambiguity in NL questions, question rewriting has emerged as an effective strategy with several key advantages: (1) \textbf{Improved Accuracy.} Rewriting ambiguous questions to better match the structure and conventions of the database helps ensure that the produced SQL accurately reflects the user's intent, thereby reducing errors in SQL generation. (2) \textbf{Enhanced Efficiency.} Clearer and more specific NL queries lead to more efficient database operations, since SQL statements can align more directly with required data retrieval or subsequent downstream tasks. Meanwhile, several challenges related to question rewriting methods need to be properly addressed. (1) \textbf{Preserving User Intent.} It’s essential that the rewritten questions maintain the original meanings. However, given the nuances of NL queries, question rewriting can sometimes alter their intended meanings, which risks generating incorrect SQL. (2) \textbf{Handling Domain-Specific Terms.} Many domains utilize specialized terminologies infrequently noticed in the LLM's training data. This lack of representation can lead to errors, since LLMs may struggle to interpret and rephrase these terms correctly, thereby generating erroneous SQL queries. To successfully rewrite questions while addressing these challenges, future research should concentrate on designing text-to-SQL systems that can accurately capture the intent behind NL queries and adapt to domain-specific terms. 

\textbf{Direction 2: Using Schema Linking Properly.} Advanced schema linking techniques is key to aligning NL questions with the correct database schema elements. Despite the prevalence of using schema linking in the existing literature \cite{dong2023c3, pourreza2024din, wang2024mac}, several critical challenges still persist: (1) \textbf{Computational Complexity}: Accurately matching NL keywords to schema elements requires substantial computational resources, especially for large databases with numerous tables and columns. Meanwhile, schema linking often results in slower response time, which is problematic for systems requiring high responsiveness. (2) \textbf{Incorrect Association with Database Schema}: Errors in schema linking can arise when relevant database information is inadvertently neglected. In particular, \cite{maamari2024death} suggested that schema linking may be unnecessary when the entire database schema fits within the LLM's context window. However, in real-world applications with large databases, text-to-SQL systems need to select only the most relevant elements rather than the whole database. As a result, how to select an optimal subset of database elements without compromising SQL accuracy and efficiency is worth investigating.

\textbf{Direction 3: Exploring more advanced FT-based approaches.} As of the time of our writing, FT-based approaches remain less common than ICL-based approaches. The reasons behind this phenomenon are three-fold: (1) \textbf{Diminished Robustness}: Fine-tuning LLMs on specific benchmark datasets often leads to reduced robustness, since these models tend to perform sub-optimally on text-to-SQL tasks with unfamiliar data. (2) \textbf{Domain-Specific Constraints}: Fine-tuning LLMs for specific domains can limit their ability to exploit the interconnections between databases, questions, and SQL queries. In scenarios demanding explicit schema-related knowledge, this constraint can lead to significant performance degradation. (3) \textbf{Higher Labeling and Computational Costs}: FT-based methods generally require substantial amounts of labeled data, thereby necessitating advanced data augmentation techniques to generate high-quality data. Meanwhile, supervised fine-tuning (SFT) on training data incurs high computational costs. As a result, more novel techniques need to be designed in the FT-based approaches to alleviate these drawbacks.

\textbf{Direction 4: Improving Model Efficiency.} The input capacity constraints and slow inference speed of LLMs highlight the need for more advanced models capable of handling larger contexts and processing information quickly. To mitigate these concerns, future research should prioritize developing precise schema filtering techniques and innovative prompt designs that reduce processing stages. In addition, architectural enhancements in FT-based approaches could accelerate model performance \cite{kou2024cllms}, which is another important research focus. For instance, techniques such as knowledge distillation \cite{cui2024distillation, zhong2024learning} and model compression \cite{ma2023llm, ramesh2023comparative} offer potential for optimizing model deployment in real-world settings. Overcoming these efficiency challenges is crucial for scaling text-to-SQL systems and supporting their broader adoption in practical scenarios.

\textbf{Direction 5: Alleviating data insufficiency.} Datasets in real-world applications are often smaller than research benchmarks, and their SQL queries tend to be more complex. Therefore, constructing diverse and accurate text-to-SQL datasets to enhance LLMs is critical \cite{long2024llms}. However, manual annotation of large datasets is costly and time-consuming, posing a significant barrier to optimizing model performance. To address data insufficiency, developing automated or semi-automated data augmentation techniques for generating additional (NL, SQL) pairs is crucial \cite{li2024codes}. Another promising solution is transfer learning \cite{patidar2024few}, which fine-tunes a model pre-trained on a large dataset using a smaller and task-specific dataset, enabling the model to capture subtleties in the new task. Additionally, federated learning algorithms \cite{zhang2023federated} can be employed to mitigate data scarcity by allowing models to benefit from data aggregated across multiple clients without compromising data privacy.

\textbf{Direction 6: Enhancing Data Security and Model Reliability.} Utilizing ICL-based methods in text-to-SQL systems raises data privacy concerns, since closed-source APIs accessing local databases can risk data leakage, particularly when sensitive information is transmitted over networks. Accordingly, the trade-off between model performance and data security emphasizes the need for customized frameworks in text-to-SQL that effectively balance both aspects. A promising approach to alleviating data privacy concerns involves exploring the synergy between open-source and proprietary LLMs \cite{li2024sea}, leveraging their respective advantages for secure and efficient SQL generation. Moreover, federated learning \cite{zhang2023federated} is another research direction for data security. Concurrently, improving model interpretability is vital for reliability \cite{ali2023explainable, zhao2024explainability}. Understanding how the model maps NL queries to database elements can foster the advancement of more robust text-to-SQL systems. For instance, approaches like explainable AI (XAI) \cite{ali2023explainable} can clarify which parts of an input query correspond to specific SQL components, providing insights into the model's internal logic. Additionally, there are some articles regarding the design of user-friendly interfaces for SQL generation \cite{zhou2024db1, tang2024mavidsql, tian2024sqlucid, rai2024understanding}, which can also empower non-technical users to comprehend and refine SQL queries more effectively. 

\textbf{Direction 7: Improving Model Robustness.} Enhancing model robustness is essential for LLM-based text-to-SQL systems, since it directly influences the model's capability to accurately interpret and convert NL queries into SQL across various scenarios. Therefore, several directions may be explored to strengthen model robustness. (1) \textbf{Developing Feedback Mechanisms.} Incorporating feedback mechanisms can significantly enhance model robustness in SQL generation. These mechanisms allow LLMs to self-evaluate the generated SQL, identify potential issues, and iteratively refine outputs \cite{li2024sea}. However, refining SQL queries based on feedback (especially SQL execution results) adds computational complexity, thereby complicating real-time or large-scale applications. As a result, developing novel feedback mechanisms to optimize model accuracy and efficiency remains a key research direction in SQL generation. (2) \textbf{Developing Output Consistency-based Techniques.} Output consistency-based approaches such as self-consistency \cite{gao2023text, shen2024selectsql} and cross-consistency \cite{li2024pet} contribute to robust SQL generation. However, higher temperatures applied to introduce randomness in self-consistency may degrade model performance, and the diversity within a single LLM may be insufficient \cite{renze2024effect}. Additionally, implementing cross-consistency requires consistency across different LLMs, which can be computationally intensive. Consequently, exploring ways to leverage LLMs for developing innovative output consistency-based methods remains an important research direction. (3) \textbf{Developing Novel Prompt Engineering Techniques.} Current prompt engineering techniques in SQL generation commonly fail to adapt well to changing data distributions or complex query requirements. Accordingly, using novel prompt engineering techniques to further enhance model robustness is another valuable research direction.

\textbf{Direction 8: Exploring Multilingual and Context-Dependent Text-to-SQL Tasks.} Multilingual SQL generation research enables the extraction of structured data from NL queries across diverse languages, facilitating efficient information access from a global perspective. Although a few multilingual text-to-SQL datasets have been introduced in the literature \cite{dou2023multispider, min2019pilot, nooralahzadeh2024statbot}, they have received limited research attention. A primary challenge lies in the diverse syntactic and semantic structures across languages, thereby complicating the parsing of NL queries into standardized SQL. Moreover, context-dependent text-to-SQL research focuses on processing multi-turn dialogues to generate SQL. This task aligns more closely with real-world applications, since users often require multiple turns to clarify ambiguous questions to produce correct SQL. Nevertheless, there are only three LLM-based studies investigating this field \cite{zhang2024coe, sun2024qda, zhang2024se}. Some primary challenges include: (1) \textbf{Context Tracking and Memory Management.} The model needs to accurately manage information from previous turns, which requires robust context tracking and memory management capabilities. (2) \textbf{Ambiguity Resolution.} Users may employ pronouns, abbreviations, or vague expressions to refer to entities mentioned previously. The model should resolve these ambiguities based on prior context, which is challenging given the nuanced nature of NL. Consequently, these tasks offer valuable potential for future research.

\begin{tcolorbox}[colback=black!5!white, colframe=black, 
                  title=\textcolor{black}{\textbf{Answer to RQ4}}, 
                  coltitle=white, colbacktitle=pink, 
                  boxrule=1pt, width=\textwidth]
(1) Existing LLM-based text-to-SQL research faces seven unsolved challenges, primarily concerning model development and data augmentation. Specifically, challenges in model development include prevalent question ambiguity, inadequate schema understanding, limited cross-domain generalization, low model efficiency, potential data security and model reliability issues along with other technical issues. Data augmentation challenges, on the other hand, stem from insufficient and low-quality data.

(2) Drawing on insights from the surveyed literature, we proposed eight future directions for advancing LLM-based text-to-SQL systems. These directions include alleviating question ambiguity and data insufficiency, leveraging schema linking effectively, exploring advanced FT-based methods, improving model efficiency, enhancing data security and model reliability, strengthening model robustness, and emphasizing multilingual and context-dependent text-to-SQL research. 
\end{tcolorbox}

\section{Threats to Validity} \label{threat}
In this survey, we anticipate to offer a systematic review of research related to LLM-based text-to-SQL. Nonetheless, two threats to validity, including \textit{Article Selection} and \textit{Data Analysis}, cannot be entirely eliminated.

\subsection{Article Selection Validity}
Article selection involves identifying all pertinent articles for the research, which is essential for establishing the theoretical framework and providing context for the survey. In this study, we meticulously selected the keywords ``\textit{Text-to-SQL}'', ``\textit{Text2SQL}'', ``\textit{Natural Language-to-SQL}'', ``\textit{NL2SQL}'', and ``\textit{SQL Generation}'' to gather articles from several widely used digital libraries. To minimize research bias, the article selection process was conducted collaboratively. The first two authors adhered strictly to the procedures and applied the inclusion and exclusion criteria outlined in Section \ref{method} for relevant article collection. Meanwhile, the remaining authors independently reviewed the article selection process to ensure the relevance of collected papers. Despite these efforts, there remains a moderate risk that some relevant studies may have been overlooked.

\subsection{Data Analysis Validity}
The threat to data analysis primarily involves extracting incomplete data items from the surveyed papers and misclassifying papers for the research questions. To minimize the risk of data item extraction, the authors in this study were organized into two groups. One group independently extracted data items from the collected articles, while the other group cross-checked the extracted data items. Any arising issues during this process were further discussed and resolved. Moreover, there is a possibility of misunderstanding LLM-based SQL generation techniques, which may result in the misclassification of relevant papers. To alleviate this, we carefully analyzed and summarized the main contributions of each paper to ensure an accurate understanding of these articles. Meanwhile, multiple authors were engaged in discussions about the classification of the surveyed articles and the survey writing process.

\section{Conclusion} \label{conclusion}
This survey presents a systematic review of the application of large language models (LLMs) in text-to-SQL tasks. We began by collecting relevant literature from multiple digital libraries, employing targeted keywords and a defined time range. After manually filtering out studies that did not meet our established inclusion and exclusion criteria, key data items were extracted to address the given research questions systematically.

In RQ1, we analyzed research trends by examining the publication dates, venues, and main contributions of the collected studies. In RQ2, we introduced the typical procedures in generating SQL queries, discussing each stage from pre-processing to post-processing in detail. In RQ3, we reviewed the existing open-source datasets and evaluation metrics for training and assessing text-to-SQL systems. In RQ4, we discussed the remaining challenges and potential future directions for LLM-based text-to-SQL, highlighting key areas for future research. This survey aims to enrich readers' understanding of recent innovations in employing LLMs in SQL generation and to inspire ongoing exploration in this rapidly evolving field.


\bibliographystyle{plain}
\bibliography{reference}

\end{document}